\documentclass[runningheads]{llncs}
\usepackage[T1]{fontenc}
\usepackage{graphicx}
\usepackage{hyperref}
\usepackage{url}
\usepackage{caption}
\usepackage{subcaption}
\usepackage{wrapfig}
\usepackage{booktabs} 
\usepackage{multirow} 
\usepackage{colortbl}
\usepackage{enumitem}
\usepackage{makecell}
\usepackage{amssymb}

\renewcommand\paragraph[1]{\vspace{2ex}\noindent\textbf{\textit{#1}}\\[1mm]}

\usepackage{amsmath,amsfonts,bm}

\def\eqref#1{equation~\ref{#1}}

\def\1{\bm{1}}

\def\mA{{\bm{A}}}

\def\mC{{\bm{C}}}

\def\mH{{\bm{H}}}

\def\mK{{\bm{K}}}

\def\mO{{\bm{O}}}
\def\mP{{\bm{P}}}
\def\mQ{{\bm{Q}}}
\def\mR{{\bm{R}}}
\def\mS{{\bm{S}}}

\def\mV{{\bm{V}}}

\DeclareMathAlphabet{\mathsfit}{\encodingdefault}{\sfdefault}{m}{sl}
\SetMathAlphabet{\mathsfit}{bold}{\encodingdefault}{\sfdefault}{bx}{n}

\def\gI{{\mathcal{I}}}

\def\gT{{\mathcal{T}}}

\newcommand{\R}{\mathbb{R}}

\usepackage{xcolor}
\newcommand{\eg}{\textit{e}.\textit{g}.}
\newcommand{\ie}{\textit{i}.\textit{e}.}

\newcommand{\model}{\textsc{ProtREM}}

\newcounter{bxincomm}
\definecolor{aqua}{rgb}{0.00,0.67,0.80}

\newcounter{todocomm}

\begin{document}
\title{Retrieval-Enhanced Mutation Mastery: Augmenting Zero-Shot Prediction of Protein Language Model}
\titlerunning{ProtREM}
%
\author{Yang Tan\inst{1,2} \orcidID{0009-0004-7261-1705}
\and
Ruilin Wang \inst{2}\orcidID{0009-0004-3675-5103} 
\and
Banghao Wu\inst{1}\orcidID{0009-0002-3361-161X}
\and
Liang Hong \inst{1} \orcidID{0000-0003-0107-336X}
\and
Bingxin Zhou \inst{1}\orcidID{0000-0002-3897-9766}}
\authorrunning{Tan et al.}
%
\institute{Shanghai Jiao Tong University, 200240, Shanghai, China
\email{\{wubh20,hong3liang,bingxin.zhou\}@sjtu.edu.cn}
\and
East China University of Science and Technology, 200231, Shanghai, China
\email{\{tyang,wrl61\}@mail.ecust.edu.cn}
}
\maketitle

\begin{abstract}
Enzyme engineering enables the modification of wild-type proteins to meet industrial and research demands by enhancing catalytic activity, stability, binding affinities, and other properties. The emergence of deep learning methods for protein modeling has demonstrated superior results at lower costs compared to traditional approaches such as directed evolution and rational design. In mutation effect prediction, the key to pre-training deep learning models lies in accurately interpreting the complex relationships among protein sequence, structure, and function. This study introduces a retrieval-enhanced protein language model for comprehensive analysis of native properties from sequence and local structural interactions, as well as evolutionary properties from retrieved homologous sequences. The state-of-the-art performance of the proposed \model~is validated on over $2$ million mutants across $217$ assays from an open benchmark (ProteinGym). We also conducted post-hoc analyses of the model’s ability to improve the stability and binding affinity of a VHH antibody. Additionally, we designed $10$ new mutants on a DNA polymerase and conducted wet-lab experiments to evaluate their enhanced activity at higher temperatures. Both \textit{in silico} and experimental evaluations confirmed that our method provides reliable predictions of mutation effects, offering an auxiliary tool for biologists aiming to evolve existing enzymes. The implementation is publicly available at \url{https://github.com/tyang816/ProtREM}.
\end{abstract}

\newpage
\section{Introduction}
Enzymes are fundamental components of synthetic biology systems. Wild-type enzymes often face limitations such as low catalytic activity, poor stability, and insufficient binding affinity, which restrict their applications in both academic research and industrial practice. Through enzyme engineering, wild-type proteins can be modified to enhance these properties, enabling them to meet the requirements of specific applications \cite{iovino2024domain_embed_search,lu2022machine,zhao2024lown_fitness,zhou2024protlgn}.

With the rapid expansion of protein databases and continuous advancements in artificial intelligence, deep learning methods offer new possibilities for enzyme engineering. Typically, a deep neural network for proteins is pre-trained on large-scale protein sequence data (with optional structural information) to learn to extract numerical representations of proteins, where the resulting embeddings are then used to score and rank candidate mutants \cite{lin2023esm2,meier2021esm1v,notin2022trancepteve}. This prediction task, known as \emph{mutation effect prediction}, is often evaluated using high-throughput datasets from deep mutation scanning (DMS), \eg, ProteinGym \cite{notin2024proteingym}.

Existing pre-training schemes can be broadly divided into three categories: sequence-based, structure-based, and evolution-based approaches. Sequence-based methods are the most popular choice. They analyze the implicit pairwise relationships between amino acids (AAs) to learn protein sequence representations. The objective of pre-training such a protein language model is to predict unknown AA types in a sequence from partial input, typically by autoregressively generating AAs along the sequence index \cite{madani2023progen,notin2022tranception} or by randomly masking a set of AAs and recovering them \cite{li2024prosst,rives2021esm1b}. The second category of structure-based methods incorporates topological inductive biases to capture stronger interactions between spatially adjacent AAs \cite{hsu2022esm-if1,zhou2024protlgn}. Geometric deep learning methods are typically applied to embed these associated structural constraints. The third category introduces multiple sequence alignment (MSA) data into the model to provide evolutionary information about proteins \cite{notin2022trancepteve,rao2021msa}. MSA reflects protein conservation and the variation patterns occurring through natural evolution. Although lacking explicit functional labels, homologous sequences offer valuable information beyond a single sequence or structure for predicting protein assays.

\begin{figure}[t]
    \centering
    \includegraphics[width=\linewidth]{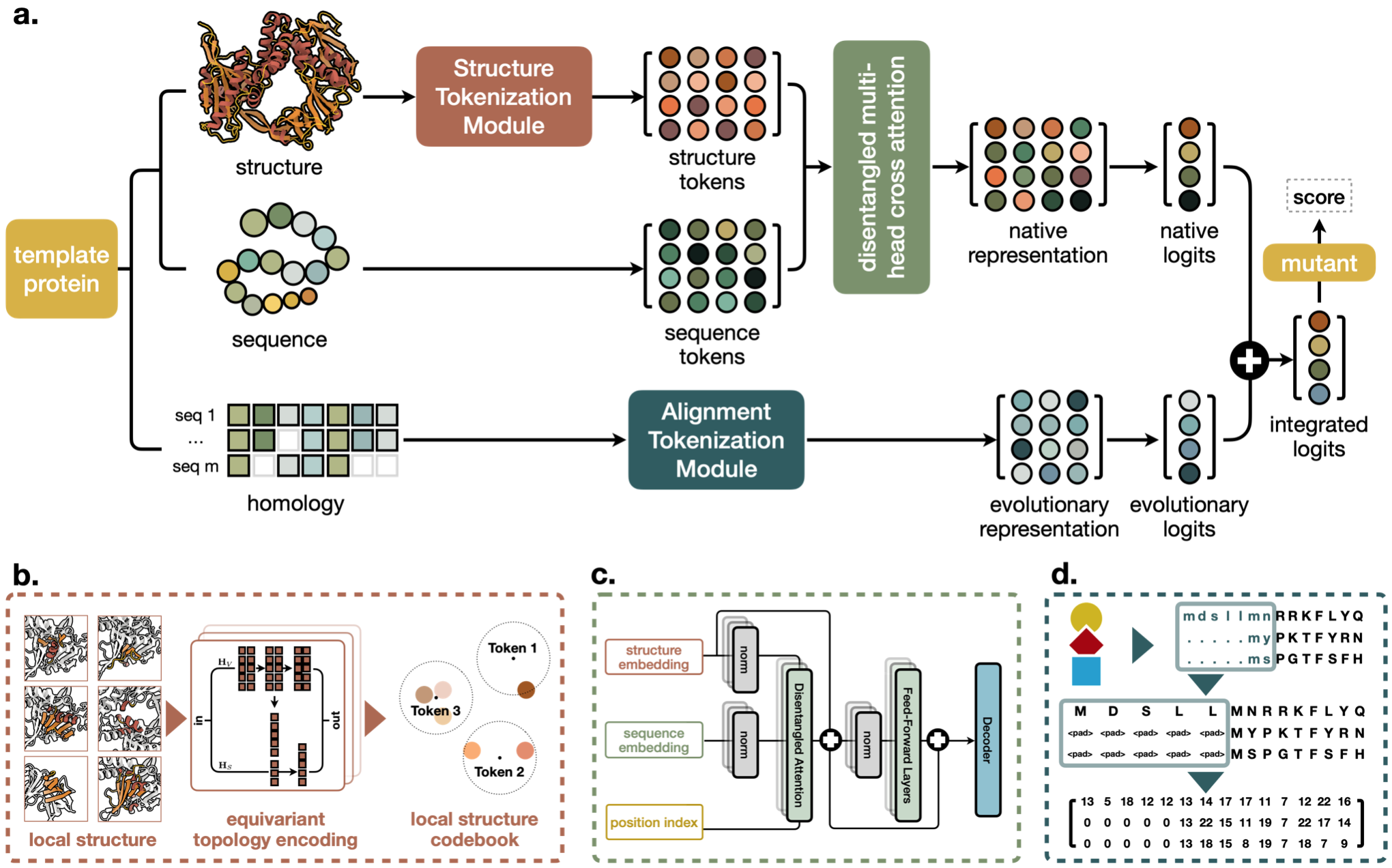}
    \caption{An illustrative workflow of \model~for predicting mutation effects. \textbf{a.} For a given template protein, \model~encodes structural, sequence, and MSA information to generate logits for each residue, which are used to calculate mutation fitness scores. \textbf{b.} For each AA, its local structure is clustered into $2048$ distinct structure tokens. \textbf{c.} The vector representations of structural and sequence information are integrated using disentangled cross-attention through BERT-style pre-training. \textbf{d.} Homologous information is retrieved via \texttt{Jackhmmer} and converted to a matrix representation of evolutionary logits.}
    \label{fig:architecture}
\end{figure}

A natural question that arises when constructing implicit representations for proteins using deep learning is: \textbf{\textit{how can we effectively integrate the non-orthogonal sequence, structure, and evolutionary characteristics?}} Although some works have considered encoding two types of information simultaneously, such as sequence and MSA information \cite{rao2021msa} or sequence and structure information \cite{tan2024protsolm,tan2023protssn,mifst}, to the best of our knowledge, no work has yet integrated all three types. Consequently, protein representations learned from pre-training frameworks may lack crucial features that could significantly impact function. To address this gap, we propose a new pre-trained protein language model with a \underline{\textbf{R}}etrieval-\underline{\textbf{E}}nhanced \underline{\textbf{M}}odule (\model). As shown in Fig.~\ref{fig:architecture}, we deliberately design the model to uncover implicit representations of native features based on sequence and structure tokens, where sequence-structure embeddings are integrated through disentangled multi-head cross-attention layers trained in a BERT-style manner. The evolutionary representations are determined by an alignment tokenization module and are integrated into the fitness evaluation of mutants.

To validate the effectiveness of the proposed \model, we examine two scenarios: (1) \textbf{\textit{Can \model~consistently outperform the prediction performance of previous methods across diverse proteins and assays on high-throughput experimental data}} (which is more comprehensive and contains more records)? (2) \textbf{\textit{Can \model~provide reliable predictions to help biologists identify expected single-site and multi-site mutation designs in practice}} (\ie, low-throughput experiments with more precise quantitative measurements)? To address the first question, we evaluate Spearman's $\rho$ between predicted mutation fitness and experimental scores on over $2$ million mutants from $217$ assays in \textbf{ProteinGym}, the largest open benchmark, and compare \model~with all models on the public leaderboard. The superior performance of our model across all functions and properties underscores its leading position in general-purpose mutation effect prediction. For the second question, we conduct both \textit{in-silico} analysis on the variable domain of the heavy chain of a nano-antibody targeting growth hormone (VHH antibody) and experimental evaluation on the bacteriophage phi29 DNA polymerase (phi29 DNAP). We discuss the model's feasibility in enhancing various assays (\eg, stability, activity, and binding affinity) with both single-site and multi-sites mutations. \model~effectively identify positive mutations and rank mutant performance scores, demonstrating significant potential for identifying highly active and stable mutants from the complete search space, thereby providing pivotal support for enhancing properties in enzyme engineering.

\section{Results}
This section presents \textit{in-silico} analyses and experimental assessments to comprehensively evaluate the performance of \model~across various properties and protein types. The following content highlights \model's performance on $217$ DMS protein assays with over $2$ million records. We also extract low-throughput experimental results from previous studies to assess the model's reliability in predicting multi-site mutants for specific proteins. Additionally, we used \model~to engineer phi29 DNAP and experimentally validated the activity and thermostability on $10$ single-site mutants.

\paragraph{\model~Ranks Top on ProteinGym Leaderboard with Significant Performance Improvement}
The first experiment assesses the model's performance on high-throughput open benchmarks. We conducted predictive evaluations using the testing data and procedures provided in the official ProteinGym repository \footnote{Available at \url{https://github.com/OATML-Markslab/ProteinGym}.}. For each of the 217 assays, we used the sequence, structure (predicted by \texttt{ColabFold 1.5} \cite{mirdita2022colabfold}), and homology sequences (retrieved by \texttt{EVCouplings} \cite{hopf2014evcouplings}) provided in the repository. For each DMS dataset, the model predicts the fitness scores of the mutants and calculates the weighted average Spearman's $\rho$ correlation between the predicted and ground truth scores. The standard deviation evaluates the stability of the model, which was computed using the bootstrap method defined in the official test script. We compared the overall performance of \model~with the top $15$ models on the leaderboard \footnote{For reliability considerations we do not include the two unreviewed models in PR.}. To avoid repetition, we retained only the highest-scoring version for each method and reported their ranks in the full leaderboard in the first column of the table. The complete leaderboard is publicly accessible at \url{https://proteingym.org/benchmarks} (26-Oct-2024). We provide the detailed configuration of \model~in Appendix~\ref{sec:app:config}. 

\begin{figure}[th]
    \centering
    \includegraphics[width=\linewidth]{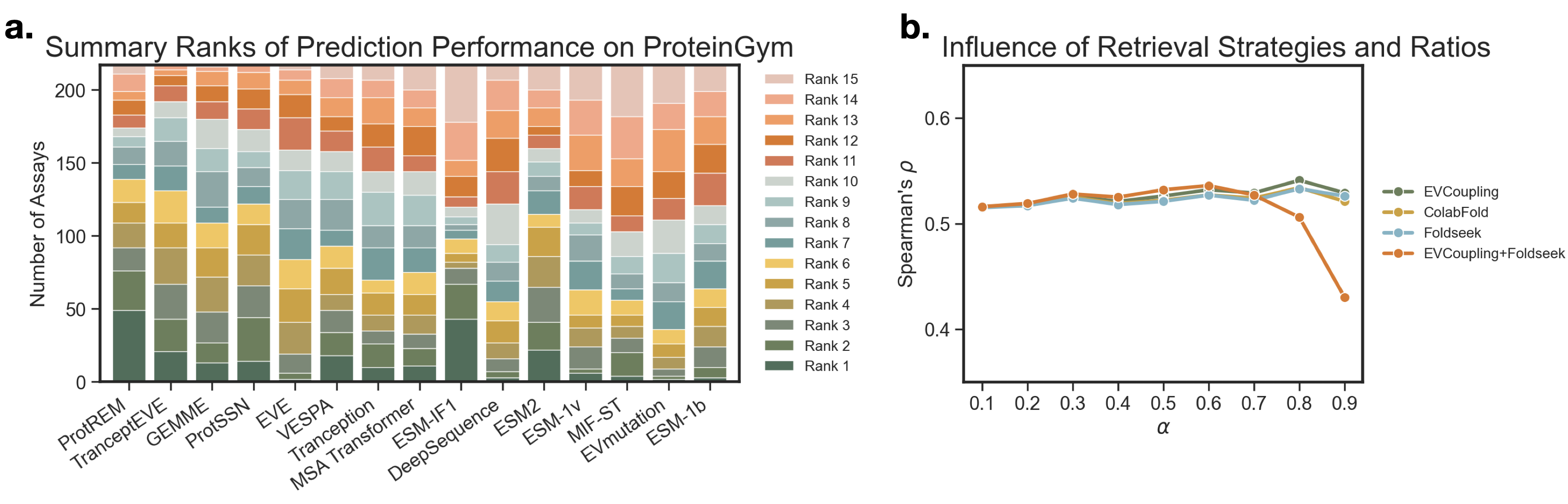}
    \caption{A summary of baseline comparisons on the ProteinGym mutation effect prediction task. \textbf{a.} Performance ranks across each assay. for instance, a Rank $1$ (in dark green) for \model~with a value of $49$ indicates that \model~achieves the highest performance on $49$ out of $217$ assays. \textbf{b.} Performance of \model's ablation models with various homologous sequence search strategies and retrieval ratios, assessed on a $10\%$ randomly split validation set.}
    \label{fig:proteingym}
\end{figure}

As reported in Table~\ref{tab:proteinGym} and Table~\ref{tab:msa_taxon_performance}, our method demonstrates significant improvements over baseline methods in both overall and property-specific evaluations. Previous studies \cite{notin2024proteingym,tan2023protssn} found that structure-aware models (\eg, SaProt, ProtSSN, and ESM-if1) are considered to perform better in binding and stability predictions, while evolution-aware models (\eg, TranceptEVE, MSA Transformer) tend to achieve higher scores in activity prediction. In response, \model~integrates sequence, structure, and evolutionary information to deliver the best overall performance and consistently ranks at the top across different individual property types. A detailed comparison of scores on individual assays is visualized in Figs.~\ref{fig:assay_activity}-\ref{fig:assay_stability}, and a summary ranking of individual predictions among all baseline methods is shown in Fig.~\ref{fig:proteingym}(a). Compared with other baseline methods, our approach has the most quantity of top-ranked assays (\eg, top 3, shown by the green bars) and the fewest lower-ranked assays (\eg, ranks after $10$, shown by the red bars). We also examined \model's performance with different homology sequence retrieval strategies (Fig.\ref{fig:proteingym}(b) and Tables~\ref{tab:retrieval_ratios}). The validation scores were assessed on a subset of ProteinGym with $10\%$ random samples from each of the $217$ assays. Overall, incorporating either sequence or structural homolog sequences improves the model's fitness prediction performance, and they are considerably insensitive to the choice of retrieval ratio (defined in (\ref{eq:alpha})), with validation scores remaining stable between $0.51$ and $0.54$. The highest performance for all methods occurs at ratios of $0.8$ and $0.6$. Based on the ablation study results on the validation set, we set the default retrieval method for \model~to \texttt{EVCouplings}, with a retrieval ratio of $0.8$.

\begin{table*}[!t]
\caption{Spearman's $\rho$ correlation of mutation effect prediction (substitution) by zero-shot predictions on \textbf{ProteinGym} of different functions.}
\label{tab:proteinGym}
\begin{center}
\resizebox{\textwidth}{!}{
    \begin{tabular}{ccccccccccc}
    \toprule
    rank & Model & seq & str & evo & \textbf{Avg. Spearman} & \textbf{Activity} & \textbf{Binding} & \textbf{Expression} & \textbf{Organismal} & \textbf{Stability} \\
    \midrule
    & \model & \checkmark & \checkmark & \checkmark & \textcolor{red}{\textbf{0.518$\pm$0.000}} & \textcolor{red}{\textbf{0.499}} & \textcolor{red}{\textbf{0.454}} & \textcolor{red}{\textbf{0.533}} & \textcolor{violet}{\textbf{0.455}} & \textcolor{red}{\textbf{0.649}} \\
    \midrule
    1 & SaProt \cite{su2023saprot} & \checkmark & \checkmark &  & \textcolor{violet}{\textbf{0.457$\pm$0.000}} & 0.458 & 0.378 & \textcolor{violet}{\textbf{0.488}} & 0.367 & \textbf{0.592} \\
    2 & TranceptEVE \cite{notin2022trancepteve} & \checkmark &  & \checkmark & \textbf{0.456$\pm$0.008} & \textcolor{violet}{\textbf{0.487}} & 0.376 & \textbf{0.457} & \textcolor{red}{\textbf{0.460}} & 0.500 \\
    4 & GEMME \cite{laine2019gemme} &  &  & \checkmark & 0.455$\pm$0.011 & \textbf{0.482} & 0.383 & 0.438 & \textbf{0.452} & 0.519 \\
    6 & ProtSSN \cite{tan2023protssn} & \checkmark & \checkmark &  & 0.449$\pm$0.006 & 0.466 & 0.366 & 0.449 & 0.396 & 0.568 \\
    10 & EVE \cite{frazer2021eve} &  &  & \checkmark & 0.439$\pm$0.010 & 0.464 & \textbf{0.386} & 0.408 & 0.447 & 0.491 \\
    14 & VESPA \cite{marquet2022vespa} & \checkmark &  &  & 0.436$\pm$0.007 & 0.468 & 0.366 & 0.404 & 0.440 & 0.500 \\
    15 & Tranception \cite{notin2022tranception} & \checkmark &  & \checkmark & 0.434$\pm$0.008 & 0.465 & 0.349 & 0.450 & 0.436 & 0.471 \\
    16 & MSA Trans \cite{rao2021msa} & \checkmark &  & \checkmark & 0.434$\pm$0.008 & 0.469 & 0.337 & 0.446 & 0.421 & 0.495 \\
    22 & ESM-if1 \cite{hsu2022esm-if1} &  & \checkmark &  & 0.422$\pm$0.011 & 0.368 & \textcolor{violet}{\textbf{0.389}} & 0.407 & 0.324 & \textcolor{violet}{\textbf{0.624}} \\
    24 & DeepSequence \cite{riesselman2018deepsequence} &  &  & \checkmark & 0.419$\pm$0.012 & 0.455 & 0.363 & 0.390 & 0.413 & 0.476 \\
    26 & ESM2 \cite{lin2023esm2} & \checkmark &  &  & 0.414$\pm$0.011 & 0.425 & 0.337 & 0.415 & 0.369 & 0.523 \\
    28 & ESM-1v \cite{meier2021esm1v} & \checkmark &  &  & 0.407$\pm$0.012 & 0.420 & 0.320 & 0.429 & 0.387 & 0.477 \\
    32 & MIF-ST \cite{mifst} & \checkmark & \checkmark &  & 0.400$\pm$0.009 & 0.390 & 0.321 & 0.438 & 0.366 & 0.485 \\
    33 & EVmutation \cite{hopf2017evmutation} &  &  & \checkmark & 0.395$\pm$0.008 & 0.440 & 0.317 & 0.378 & 0.411 & 0.430 \\
    34 & ESM-1b \cite{rives2021esm1b}  & \checkmark &  &  & 0.394$\pm$0.009 & 0.428 & 0.287 & 0.406 & 0.351 & 0.500 \\
    \bottomrule\\[-2.5mm]
    \multicolumn{10}{l}{$\dagger$ The top three scores are highlighted by \textbf{\textcolor{red}{First}}, \textbf{\textcolor{violet}{Second}}, and \textbf{Third}.}
    \end{tabular}
}
\vspace{-10mm}
\end{center}
\end{table*}

\begin{figure}[t]
    \centering
    \includegraphics[width=\linewidth]{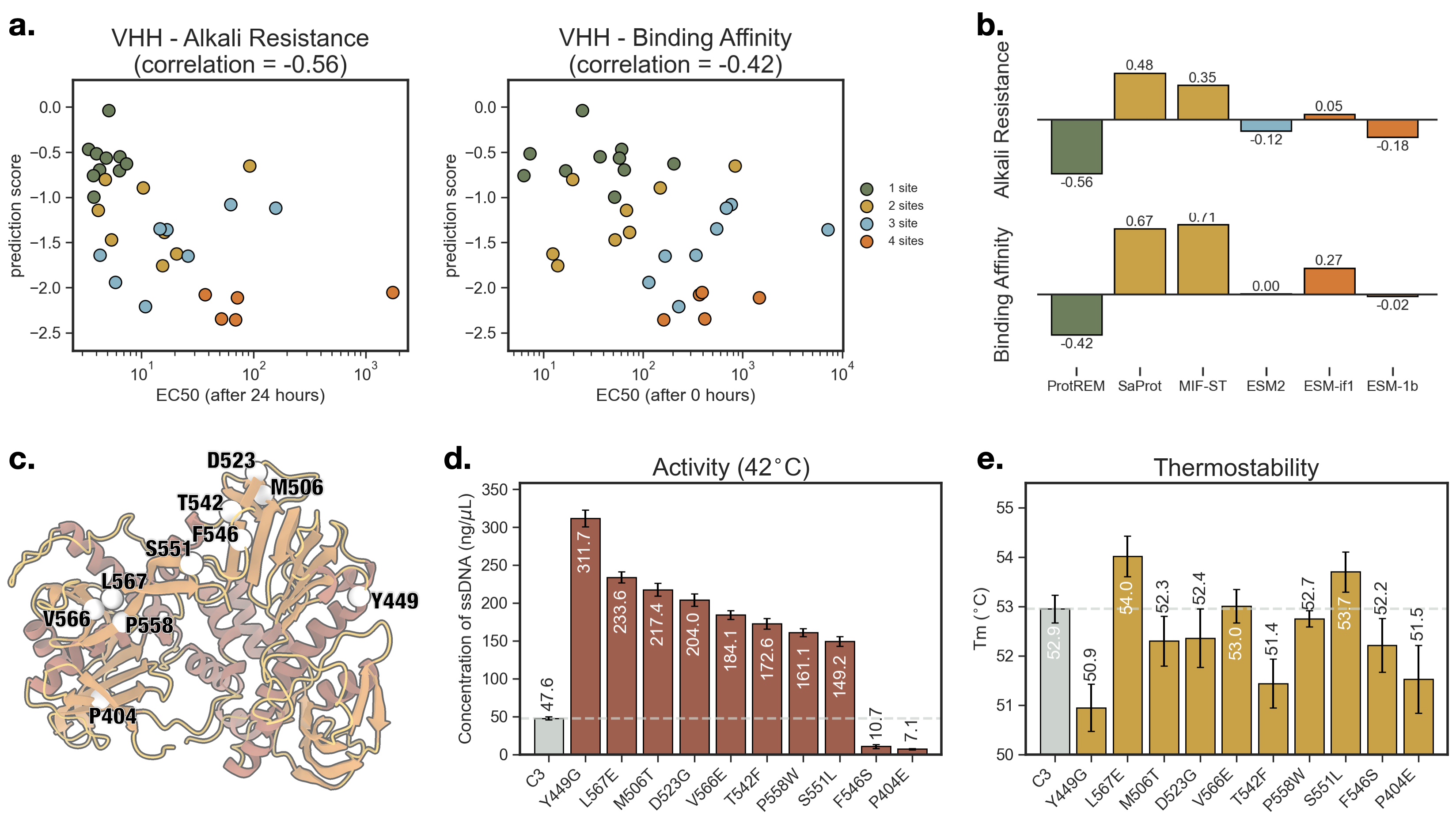}
    \caption{Performance analysis on low-throughput experimental datasets. (a) Scatter plot of predicted fitness scores (by \model) versus experimentally obtained EC50 values. For both alkali resistance and binding affinity improvements, \model’s scoring of $31$ VHH antibody mutants by 1-4 sites shows a clear correlation with experimental data. (b) Performance of different models on the two assays of VHH antibody data. Only \model~successfully generated fitness scores that are moderately negatively correlated with EC50 values. (c) 3D structure of the template phi29 DNAP. The AA sites targeted for mutation across the 10 single-site mutants are highlighted and labeled with their wild-type residues. (d) Activity improvements in phi29 DNAP mutants. Among the $10$ single-site mutants experimentally tested, $8$ shows significant activity enhancements, with the top mutant exhibiting an $8$-fold increase. (e) Thermostability of phi29 DNAP mutants. Three mutants demonstrate improvements in both thermostability and activity, with two of them showing significant gains.}
    \label{fig:caseStudy}
\end{figure}

\paragraph{\model~Performs Reliable Prediction on Favorable Alkali Resistance and Binding Affinity of VHH Nanoantibody Mutants}
The second evaluation validates \model's effectiveness in fitness prediction for both single-site and multi-site mutants. The engineering target is to enhance binding affinity and alkali resistance in the VHH antibody. This antibody type plays a crucial role in the development of clinical antibody-based therapeutics, serving as an affinity ligand to selectively purify biopharmaceuticals \cite{wang2014heterologous}.
In production, a CIP (clean-in-place) process is typically required, which involves cleaning with an extremely alkaline solution for $24$ hours to eliminate impurities and contaminants. Therefore, improving the binding affinity and alkali resistance of the VHH antibody is a key requirement for industrial production. Here, we used experimental data from \cite{kang2024vhh} on $31$ VHH antibody mutants with modifications at $1-4$ sites. Their binding affinity and alkali resistance were assessed by EC50 values before and after treatment with $0.5$ M NaOH for $24$ hours. 

Fig.~\ref{fig:caseStudy}(a) visualizes \model's fitness prediction scores with respect to experimental results (retrieved from \cite{kang2024vhh}). Here, each point represents a mutant, and the gray pentagon denotes the wild-type template. Mutants are color-coded based on mutation depth. Overall, there is a significant correlation between the fitness prediction scores and experimental results. Note that higher EC50 values indicate poorer binding affinity and alkali resistance for VHH. Thus, a correlation closer to $-1$ represents better fitness prediction. In Fig.~\ref{fig:caseStudy}(b), we also compare \model~with other popular sequence- and structure-based baseline methods with their top-performing versions from ProteinGym. Results indicate that most models fail to establish a negative correlation between fitness and the two experimental indicators. Although ESM2 and ESM1b achieve a negative correlation, the values remain close to zero, suggesting limited practical utility for these models in engineering the VHH antibody.

\paragraph{\model~Engineers phi29 DNAP Toward Significant Activity Enhancement with High Positive Rate}
In the third experiment, we used \model~to enhance the catalytic activity of phi29 DNAP at higher temperatures. phi29 DNAP plays a key role in advancing isothermal amplification methods such as multiple displacement amplification (MDA) and rolling-circle amplification (RCA) \cite{povilaitis2016vitro}. It facilitates efficient DNA amplification with low error rates, due to its remarkable strand-displacing activity with high fidelity and transcription processivity. While the existing phi29 DNAP exhibits high processivity and robust strand displacement activity at 30$^{\circ}$C, its activity and stability decline significantly at elevated temperatures. This limitation restricts its applications, particularly in cases requiring specific thermal conditions to optimize specificity or to denature complex secondary structures in DNA templates \cite{kamtekar2004insights,manrao2012reading,de2010improvement}.

Therefore, we employed \model~to enhance the activity of phi29 DNAP at a higher temperature (\ie, 42$^{\circ}$C). We selected the C3 mutant (Table~\ref{tab:phi29sequences}) as the template for further improvement, which is a phi29 DNAP mutant obtained through earlier directed evolution with high thermostability. However, it exhibits low ssDNA yield during \textit{in vitro} amplification at 42$^{\circ}$C (Fig.~\ref{fig:caseStudy}(d)) and requires further enhancement \cite{povilaitis2016vitro}. We designed $10$ single-site mutations from C3 and evaluated their RCA activity at 42$^{\circ}$C using an ssDNA template (Fig.~\ref{fig:caseStudy}(c)). The average results from $3$ independent repetitions show that $8$ of these mutants exhibit a significant activity enhancement over C3 with an over $3$-fold increase (Fig.~\ref{fig:caseStudy}(d)). The top-performing mutant, phi29 DNAP$^{\rm Y449G}$, achieved a $6.5$-fold increase in activity. Additionally, we assessed the thermostability of the mutants using differential scanning fluorimetry (DSF), observing a clear increase in T$_{\rm m}$ values in $3$ of the $8$ active positive mutants (Fig.~\ref{fig:caseStudy}(e)). Notably, in directed evolution, there is generally a trade-off between activity and stability. However, the phi29 DNAP mutants obtained by \model, such as phi29 DNAP$^{\rm L567E}$ and phi29 DNAP$^{\rm S551L}$, demonstrate simultaneous improvements in both assays, presenting a new possibility for designing high-performance variants.

\section{Methods}
This section overviews the construction of our proposed \model, the pipeline for mutation effect prediction, and the experimental methods used to assess the new mutants for Phi29. Additional details, such as the data processing for mutants in the case studies, can be found in Appendices.

\paragraph{Overview of \model}
The core of \model~consists of two main components. The first is a pre-trained PLM, which is used to find a native representation of the input single protein based on its sequence and structure (Fig.~\ref{fig:architecture}(a)). The sequence input is naturally discrete and straightforward to tokenize. For the local structure of AAs, we employ a \emph{structure tokenization module} to create a codebook. It is then combined with sequence-level tokens using a \emph{disentangled cross-attention module}. The other module extracts \emph{evolutionary representations} by searching and aligning homologous family sequences of the template protein. Both native representation and evolutionary representations are vector descriptions of AA positions and types. They are processed through an explicitly defined scoring rule to compute the final mutation fitness score.

\paragraph{Structure Tokenization Module}
The process of finding token representations for local protein structures involves two key steps, including local structure extraction and latent structure vocabulary assignment (Fig.~\ref{fig:architecture}(b)). Local structures are defined on a per-residue basis \cite{tan2024protloca}. Consider a protein with $L$ AAs. For each residue, we define its local structure by considering up to $40$ neighboring residues within a 10\AA~spatial radius. The local structure is described by an undirected graph with each node representing a residue, and two nodes are connected if their spatial distances are less than 10\AA. The constructed $L$ graphs are then fed into pre-trained GVP \cite{gvp} layers, following mean pooling operations to produce $L$ vectors of $256$ dimensions each. These graphs are then fed into a pre-trained GVP, followed by average pooling to produce $L$ vectors of $256$ dimensions each. These $256$-dimensional vectors in the continuous space are further mapped to a $2048$-dimensional discrete space using K-means clustering, where each dimension represents an implicit vocabulary of structures. Additional details for training and inferencing can be found in Appendix~\ref{sec:app:structureToken}.

\paragraph{Disentangled Multi-Head Cross Attention}
The native representation (Fig.~\ref{fig:architecture}(c)), \ie, the joint embeddings of protein sequences and structures, are obtained by a masked language model. The core propagation rule is based on disentangled multi-head cross attention \cite{he2020deberta,li2024prosst}. For a protein of length $L$, we define the token inputs for its AA sequence and structure as $\mR$ and $\mS$. 
For two arbitrary AAs at positions $i$ and $j$, their attention score ${\rm Attn}(i,j)$ is calculated based on $\mR$, $\mS$, and their relative position $\mP$:
\begin{equation}
\label{eq:attn_init}
    {\rm Attn}(i,j)=\{\mR_i,\mS_i,\mP_{ij}\}\times\{\mR_j,\mS_j,\mP_{ji}\}^{\top},
\end{equation}
where $\mP_{ij}$ represents the relative position from the $i$th AA to the $j$th AA. After the element-wise multiplication on (\ref{eq:attn_init}), we retain only the five AA-relevant attention scores and rewrite them with the query, key, and value matrices, which are obtained via linear transformations during forward propagation:
\begin{align}
\label{eq:attn_5}
    {\rm Attn}(i,j) &= \mR_i\mR_j^{\top}+\mR_i\mS_j^{\top}+\mR_i\mP_{ji}^{\top}+\mS_i\mR_j^{\top}+\mP_{ij}\mR_j^{\top} \\
    &= \mQ_i^R(\mK_j^R)^{\top}+\mQ_i^R(\mK_j^{S})^{\top}+\mQ_i^R(\mK_{ji}^{P})^{\top}+\mQ_i^S(\mK_j^R)^{\top}+\mQ_{ij}^P(\mK_j^{R})^{\top}.\notag
\end{align}
The notations follows conventional attention. The subscription denotes their association with the AA and the superscription denotes their association to the input, \eg, $\mQ_i^R$ represents the query value for the $i$th AA on its sequence information, and $\mK_{ij}^P$ represents the key value with respect to the relative distance from the $i$th AA to the $j$th AA.

Similar to classic attention schemes, the initial attention scores obtained from (\ref{eq:attn_5}) requires element-wise normalization by a scaling factor $1/\sqrt{5d}$ with $d$ being the dimension of $\mQ^R$. Denote the unnormalized attention matrix as $\mH_{\rm attn}$. We use it to update the hidden representation for $\mR$, which reads:
\begin{equation}
    \mR_o=\sigma(\frac{\mH_{\rm attn}}{\sqrt{5d}})\mV^R,
\end{equation}
where $\sigma(\cdot)$ is a softmax activation function, and $\mV^R$ is the value matrix of $\mR$.

\paragraph{Homology Retrieval}
In addition to the sequence- and structure-based native information, \model~also incorporates evolution information. We employ \texttt{EVCouplings} \cite{hopf2014evcouplings} and filters results from \texttt{Jackhmmer} \cite{eddy2011jackhmmer} by maximizing the number of significant hits with a bit score between $0.1$ and $0.9$. The second strategy follows the retrieval method in \texttt{ColabFold} \cite{mirdita2022colabfold}. We also considered different sequence retrieval strategies and structure homologs retrieval (Appendix~\ref{sec:app:ablation}). For simplicity, we replace the gap character \texttt{-} with a special token \texttt{<pad>}. Suppose $N$ homologous sequences are found for a protein of length $L$, denoted by $\mA\in\mathbb{Z}^{N\times L}$. A counting matrix $\mC\in\R^{L\times V}$ records the frequencies of AA types at each residue position, where
\begin{equation}
    \mC_{iv}= \frac{\sum_{n=1}^N\gI(\mA_{ni}=v)}{\sum_{v=1}^V\sum_{n=1}^N\gI(\mA_{ni}=v)},
\end{equation}
where $\mA\in\mathbb{Z}^{N\times L}$. The vocabulary size $V=25$, accounting for $20$ AAs and $5$ special tokens. The indicator function $\gI(\mA_{ki}=v)$ assigns $1$ if the $i$th position ($1\leq i\leq L$) of the $n$th sequence ($1\leq n\leq N$) fills the $v$th token ($1\leq v\leq V$), otherwise $0$. The final evolutionary logits $\mO^{\rm evo}$ is generated from $\mC$ with subsequent normalization and log-transformation, \ie 
\begin{equation}
\label{eq:msaFitness}
    \mO^{\rm evo}_{iv}=\log\left(\frac{\exp(\mC_{iv})}{\sum_{v=1}^V\exp(\mathbf{C}_{iv})}\right).
\end{equation}

\paragraph{Zero-Shot Fitness Scoring}
For a given mutant, its fitness score is calculated by comparing the predicted logits of the mutant residues with those of the wild-type residues. We first combine the predicted native and evolutionary logits:
\begin{equation}
\label{eq:alpha}
    \mO^{\rm out}_{iv}=(1-\alpha)\cdot\mO^{\rm native}_{iv}+\alpha\cdot\mO^{\rm evo}_{iv},
\end{equation}
where the retrieval ratio $\alpha\in[0,1]$ controls the weight of integrating the intrinsic and evolution probability distributions. 

The mutation fitness scores are calculated from $\mO^{\rm out}$. For a mutant $x$, the overall fitness score $F_x$ is obtained by summing the associated logit differences across all mutation sites $t\in\gT$:
\begin{equation}
    F_x=\sum_{t\in\gT}\left(\mO^{\rm out}_{tv^{\prime}}-\mO^{\rm out}_{tv}\right),
\end{equation}
where at the position $t$, the mutant alters the residue type from $v$ to $v^{\prime}$. This scoring method captures the relative fitness by evaluating how much the mutation alters the predicted logit values compared to the wild type, with larger logit differences reflecting more significant deviations in fitness. For multiple mutations, the logit differences are summed over all the mutation sites, reflecting the cumulative effect of all alterations.

\paragraph{Rolling circle replication assay of phi29 DNA polymerase}
We assessed the RCA reaction to simulate the circular template amplification activity of phi29 DNA polymerase. Rolling circle amplification (RCA) was conducted using the M13mp18 single-stranded DNA (ssDNA) template (NEB, United States) in a 20 $\mu$L reaction volume. The reaction mixture included 2 $\mu$L of M13mp18 template (0.2 $\mu$g/$\mu$L), 2 $\mu$L of 10× RCA buffer (Thermo Fisher, United States), 2 $\mu$L of phi29 DNA polymerase (100 ng/$\mu$L), 4.8 $\mu$L of primer mix (2.4 $\mu$M final concentration), 0.8 $\mu$L of dNTPs (10 mM each), and 8.4 $\mu$L of DEPC-treated water (LABTOP BIO, China). The RCA was performed at 42$^{\circ}$C for 10 minutes, followed by a denaturation step at 65$^{\circ}$C for 10 minutes. DNA concentrations were subsequently measured using the ssDNA Assay Kit (YEASEN, China) with a Qubit4 fluorometer (Thermo Fisher, United States). 

\paragraph{Differential Scanning Fluorimetry (DSF)} 
The T$_{\rm m}$ values were measured using the DSF method with the Protein Thermal Shift Dye Kit (Thermo Fisher, U.S.A.). To prepare the reaction mixture, 1.0 $\mu$L of SYPRO Orange Dye (SUPELCO, U.S.A.) was added to 49 $\mu$L of lysis buffer (25 mM Tris-HCl, 500 mM NaCl, pH 7.5). Subsequently, 1 $\mu$L of the diluted dye was combined with 19 $\mu$L of a 0.2 mg/mL protein solution. DSF experiments were conducted using the LightCycler 480 Instrument II (Roche, U.S.A.). The reaction mixture was initially set at 25 $^{\circ}$C then heated gradually to 99 $^{\circ}$C at a rate of 0.1 $^{\circ}$C s-1. Data analysis was performed by Protein Thermal Shift.

\section{Discussion and Conclusion}
Protein engineering is a central topic in synthetic biology. where novel mutants are typically found through rational design, machine learning, or pre-trained deep neural networks. The latest pre-trained deep learning methods have demonstrated great advances in identifying favorable mutations. To evaluate these methods with minimum experimental costs, a standard workflow has been established to evaluate the models' scoring performance on large-scale DMS datasets. The establishment of such an evaluation standard enables direct comparison between deep learning models. It also allows researchers without biological experimental capabilities to independently assess their models.

A protein embedding algorithm could benefit from incorporating additional protein modalities to yield more expressive vector representations. For example, as demonstrated by \cite{notin2024proteingym,tan2023protssn}, structure-aware models are generally more effective at enhancing stability and binding affinity, while sequence-centric models tend to improve activity. However, the potential insights provided by evolutionary information remain less explored. For instance, relatively conserved regions in proteins are typically poor candidates to mutate, and AA  conservations are primarily drawn from evolutionary analysis on homologous sequence data. Moreover, existing structural models often rely on predicted 3D structures especially when crystallographic structures are unavailable. Since these predicted structures may contain significant inaccuracies, homologous sequences could help mitigate the issue by providing an explicit coevolutionary relationship of residue pairs. 

On the other hand, the existing evaluation framework relies on high-throughput DMS datasets. However, these experiments are prone to inaccuracies, and most entities are negative mutants. Thus, we extract low-throughput experimental data and conduct multi-dimensional post-hoc analysis, offering an approach similar to wet-lab validation without requiring new experiments. Furthermore, we select an important enzyme with clear modification demands, perform single-site mutations, and validate these through experiments. All experimental and analytical results indicate that our model significantly outperforms other methods on large-scale datasets, demonstrates higher reliability on small-scale experimental data, and identifies novel mutants with improved functions and properties in practical applications. We believe this work holds significant value for both the computer science and biology communities, from model design, method evaluation, and tool application to the future use of specific enzyme products.

\begin{credits}
\subsubsection{\ackname} This work was supported by the grants from the National Science Foundation of China (Grant Number 12104295, 62302291), the Computational Biology Key Program of Shanghai Science and Technology Commission (23JS1400600), Shanghai Jiao Tong University Scientific and Technological Innovation Funds (21X010200843), and Science and Technology Innovation Key R\&D Program of Chongqing (CSTB2022TIAD-STX0017), the Student Innovation Center at Shanghai Jiao Tong University, and Shanghai Artificial Intelligence Laboratory.

\subsubsection{\discintname}
The authors have no competing interests to declare that are relevant to the content of this article. 
\end{credits}
%
%
%
\bibliographystyle{splncs04}
\bibliography{1reference}

\begin{thebibliography}{10}
\providecommand{\url}[1]{\texttt{#1}}
\providecommand{\urlprefix}{URL }
\providecommand{\doi}[1]{https://doi.org/#1}

\bibitem{abramson2024alphafold3}
Abramson, J., Adler, J., Dunger, J., Evans, R., Green, T., Pritzel, A., Ronneberger, O., Willmore, L., Ballard, A.J., Bambrick, J., et~al.: Accurate structure prediction of biomolecular interactions with {AlphaFold} 3. Nature pp.~1--3 (2024)

\bibitem{eddy2011jackhmmer}
Eddy, S.R.: Accelerated profile hmm searches. PLoS Computational Biology  \textbf{7}(10),  e1002195 (2011)

\bibitem{frazer2021eve}
Frazer, J., Notin, P., Dias, M., Gomez, A., Min, J.K., Brock, K., Gal, Y., Marks, D.S.: Disease variant prediction with deep generative models of evolutionary data. Nature  \textbf{599}(7883),  91--95 (2021)

\bibitem{he2020deberta}
He, P., Liu, X., Gao, J., Chen, W.: Deberta: Decoding-enhanced bert with disentangled attention. arXiv:2006.03654  (2020)

\bibitem{hopf2017evmutation}
Hopf, T.A., Ingraham, J.B., Poelwijk, F.J., Sch{\"a}rfe, C.P., Springer, M., Sander, C., Marks, D.S.: Mutation effects predicted from sequence co-variation. Nature Biotechnology  \textbf{35}(2),  128--135 (2017)

\bibitem{hopf2014evcouplings}
Hopf, T.A., Sch{\"a}rfe, C.P., Rodrigues, J.P., Green, A.G., Kohlbacher, O., Sander, C., Bonvin, A.M., Marks, D.S.: Sequence co-evolution gives {3D} contacts and structures of protein complexes. eLife  \textbf{3},  e03430 (2014)

\bibitem{hsu2022esm-if1}
Hsu, C., Verkuil, R., Liu, J., Lin, Z., Hie, B., Sercu, T., Lerer, A., Rives, A.: Learning inverse folding from millions of predicted structures. In: International Conference on Machine Learning. pp. 8946--8970. PMLR (2022)

\bibitem{iovino2024domain_embed_search}
Iovino, B.G., Tang, H., Ye, Y.: Protein domain embeddings for fast and accurate similarity search. In: International Conference on Research in Computational Molecular Biology. pp. 421--424. Springer (2024)

\bibitem{gvp}
Jing, B., Eismann, S., Suriana, P., Townshend, R.J.L., Dror, R.: Learning from protein structure with geometric vector perceptrons. In: International Conference on Learning Representations (2021)

\bibitem{kamtekar2004insights}
Kamtekar, S., Berman, A.J., Wang, J., L{\'a}zaro, J.M., de~Vega, M., Blanco, L., Salas, M., Steitz, T.A.: Insights into strand displacement and processivity from the crystal structure of the protein-primed {DNA} polymerase of bacteriophage $\varphi$29. Molecular Cell  \textbf{16}(4),  609--618 (2004)

\bibitem{kang2024vhh}
Kang, L., Wu, B., Zhou, B., Tan, P., Kang, Y., Yan, Y., Zong, Y., Li, S., Liu, Z., Hong, L.: Ai-enabled alkaline-resistant evolution of protein to apply in mass production. bioRxiv pp. 2024--09 (2024)

\bibitem{laine2019gemme}
Laine, E., Karami, Y., Carbone, A.: Gemme: a simple and fast global epistatic model predicting mutational effects. Molecular Biology and Evolution  \textbf{36}(11),  2604--2619 (2019)

\bibitem{li2024prosst}
Li, M., Tan, Y., Ma, X., Zhong, B., Yu, H., Zhou, Z., Ouyang, W., Zhou, B., Hong, L., Tan, P.: {ProSST}: Protein language modeling with quantized structure and disentangled attention. bioRxiv pp. 2024--04 (2024)

\bibitem{lin2023esm2}
Lin, Z., Akin, H., Rao, R., Hie, B., Zhu, Z., Lu, W., Smetanin, N., Verkuil, R., Kabeli, O., Shmueli, Y., et~al.: Evolutionary-scale prediction of atomic-level protein structure with a language model. Science  \textbf{379}(6637),  1123--1130 (2023)

\bibitem{lu2022machine}
Lu, H., Diaz, D.J., Czarnecki, N.J., Zhu, C., Kim, W., Shroff, R., Acosta, D.J., Alexander, B.R., Cole, H.O., Zhang, Y., et~al.: Machine learning-aided engineering of hydrolases for {PET} depolymerization. Nature  \textbf{604}(7907),  662--667 (2022)

\bibitem{madani2023progen}
Madani, A., Krause, B., Greene, E.R., Subramanian, S., Mohr, B.P., Holton, J.M., Olmos, J.L., Xiong, C., Sun, Z.Z., Socher, R., et~al.: Large language models generate functional protein sequences across diverse families. Nature Biotechnology  \textbf{41}(8),  1099--1106 (2023)

\bibitem{manrao2012reading}
Manrao, E.A., Derrington, I.M., Laszlo, A.H., Langford, K.W., Hopper, M.K., Gillgren, N., Pavlenok, M., Niederweis, M., Gundlach, J.H.: Reading {DNA} at single-nucleotide resolution with a mutant mspa nanopore and phi29 {DNA} polymerase. Nature Biotechnology  \textbf{30}(4),  349--353 (2012)

\bibitem{marquet2022vespa}
Marquet, C., Heinzinger, M., Olenyi, T., Dallago, C., Erckert, K., Bernhofer, M., Nechaev, D., Rost, B.: Embeddings from protein language models predict conservation and variant effects. Human Genetics  \textbf{141}(10),  1629--1647 (2022)

\bibitem{meier2021esm1v}
Meier, J., Rao, R., Verkuil, R., Liu, J., Sercu, T., Rives, A.: Language models enable zero-shot prediction of the effects of mutations on protein function. Advances in Neural Information Processing Systems  \textbf{34},  29287--29303 (2021)

\bibitem{mirdita2022colabfold}
Mirdita, M., Sch{\"u}tze, K., Moriwaki, Y., Heo, L., Ovchinnikov, S., Steinegger, M.: Colabfold: making protein folding accessible to all. Nature Methods  \textbf{19}(6),  679--682 (2022)

\bibitem{notin2022tranception}
Notin, P., Dias, M., Frazer, J., Hurtado, J.M., Gomez, A.N., Marks, D., Gal, Y.: Tranception: protein fitness prediction with autoregressive transformers and inference-time retrieval. In: International Conference on Machine Learning. pp. 16990--17017. PMLR (2022)

\bibitem{notin2024proteingym}
Notin, P., Kollasch, A., Ritter, D., Van~Niekerk, L., Paul, S., Spinner, H., Rollins, N., Shaw, A., Orenbuch, R., Weitzman, R., et~al.: {ProteinGym}: large-scale benchmarks for protein fitness prediction and design. In: Advances in Neural Information Processing Systems. vol.~36 (2024)

\bibitem{notin2022trancepteve}
Notin, P., Van~Niekerk, L., Kollasch, A.W., Ritter, D., Gal, Y., Marks, D.S.: Trancepteve: Combining family-specific and family-agnostic models of protein sequences for improved fitness prediction. NeurIPS 2022 Workshop on Learning Meaningful Representations of Life  (2022)

\bibitem{povilaitis2016vitro}
Povilaitis, T., Alzbutas, G., Sukackaite, R., Siurkus, J., Skirgaila, R.: In vitro evolution of phi29 {DNA} polymerase using isothermal compartmentalized self replication technique. Protein Engineering, Design and Selection  \textbf{29}(12),  617--628 (2016)

\bibitem{rao2021msa}
Rao, R.M., Liu, J., Verkuil, R., Meier, J., Canny, J., Abbeel, P., Sercu, T., Rives, A.: {MSA} transformer. In: International Conference on Machine Learning. pp. 8844--8856. PMLR (2021)

\bibitem{riesselman2018deepsequence}
Riesselman, A.J., Ingraham, J.B., Marks, D.S.: Deep generative models of genetic variation capture the effects of mutations. Nature Methods  \textbf{15}(10),  816--822 (2018)

\bibitem{rives2021esm1b}
Rives, A., Meier, J., Sercu, T., Goyal, S., Lin, Z., Liu, J., Guo, D., Ott, M., Zitnick, C.L., Ma, J., et~al.: Biological structure and function emerge from scaling unsupervised learning to 250 million protein sequences. Proceedings of the National Academy of Sciences  \textbf{118}(15),  e2016239118 (2021)

\bibitem{su2023saprot}
Su, J., Han, C., Zhou, Y., Shan, J., Zhou, X., Yuan, F.: {SaProt}: protein language modeling with structure-aware vocabulary. In: The Twelfth International Conference on Learning Representations (2023)

\bibitem{tan2024protsolm}
Tan, Y., Zheng, J., Hong, L., Zhou, B.: {ProtSolM}: Protein solubility prediction with multi-modal features. arXiv:2406.19744  (2024)

\bibitem{tan2024protloca}
Tan, Y., Zheng, L., Zhong, B., Hong, L., Zhou, B.: Protein representation learning with sequence information embedding: Does it always lead to a better performance? arXiv:2406.19755  (2024)

\bibitem{tan2023protssn}
Tan, Y., Zhou, B., Zheng, L., Fan, G., Hong, L.: Semantical and topological protein encoding toward enhanced bioactivity and thermostability. bioRxiv pp. 2023--12 (2023)

\bibitem{van2024foldseek}
Van~Kempen, M., Kim, S.S., Tumescheit, C., Mirdita, M., Lee, J., Gilchrist, C.L., S{\"o}ding, J., Steinegger, M.: Fast and accurate protein structure search with {Foldseek}. Nature Biotechnology  \textbf{42}(2),  243--246 (2024)

\bibitem{de2010improvement}
de~Vega, M., L{\'a}zaro, J.M., Menc{\'\i}a, M., Blanco, L., Salas, M.: Improvement of $\varphi$29 {DNA} polymerase amplification performance by fusion of {DNA} binding motifs. Proceedings of the National Academy of Sciences  \textbf{107}(38),  16506--16511 (2010)

\bibitem{wang2014heterologous}
Wang, J., Bever, C.R., Majkova, Z., Dechant, J.E., Yang, J., Gee, S.J., Xu, T., Hammock, B.D.: Heterologous antigen selection of camelid heavy chain single domain antibodies against tetrabromobisphenol a. Analytical Chemistry  \textbf{86}(16),  8296--8302 (2014)

\bibitem{mifst}
Yang, K.K., Zanichelli, N., Yeh, H.: Masked inverse folding with sequence transfer for protein representation learning. Protein Engineering, Design and Selection  \textbf{36},  gzad015 (2023)

\bibitem{zhao2024lown_fitness}
Zhao, J., Zhang, C., Luo, Y.: Contrastive fitness learning: Reprogramming protein language models for low-n learning of protein fitness landscape. In: International Conference on Research in Computational Molecular Biology. pp. 470--474. Springer (2024)

\bibitem{zhou2024protlgn}
Zhou, B., Zheng, L., Wu, B., Tan, Y., Lv, O., Yi, K., Fan, G., Hong, L.: Protein engineering with lightweight graph denoising neural networks. Journal of Chemical Information and Modeling  (2024)

\end{thebibliography}

\newpage
\appendix
\setcounter{figure}{0}
\renewcommand{\figurename}{Fig.}
\renewcommand{\thefigure}{S\arabic{figure}}
\setcounter{table}{0}
\renewcommand{\tablename}{Table.}
\renewcommand{\thetable}{S\arabic{table}}

\section{Details of \model}
\label{sec:app:model}

\subsection{Structure Tokenization}
\label{sec:app:structureToken}

\subsubsection{Structure Encoder}
The structure encoder in this study is based on geometric vector perceptrons (GVP) \cite{gvp}, represented by the function $\pi_\theta(G)\in\mathbb{R}^{l\times d}$, where $l$ is the number of nodes and $d$ is the embedding dimension. The GVP is integrated with a decoder to form an auto-encoder, trained with a denoising pre-training objective by perturbing $C_\alpha$ oordinates with 3D Gaussian noise and applying Brownian motion. After training on the \texttt{CATH43-S40} dataset, we use the mean-pooled encoder output as the final representation of local structures. The encoding process for a graph $G$ is defined as:
\begin{equation}
    r=\frac1l\sum_{i=1}^l\pi_\theta(g_i),
\end{equation}
where $g_i$ represents the local structural features for the $i$-th node, and $r\in\mathbb{R}^{d}$ is the mean-pooled output.

The dataset used for training is derived from \texttt{CATH43-S40}, containing 31,270 protein structural domains after filtering. A validation set of 200 structures was randomly selected, and the model was trained with the configuration that minimized validation loss.

\subsubsection{Local Structure Codebook}
The dataset used for training the structure codebook consists of local protein structures extracted from the \texttt{CATH43-S40} database. For each protein structure, a sliding window method is applied along the residue sequence to select segments, with a specific residue as the anchor. Up to 40 residues within a 10 Å distance from the anchor are connected to form a star-shaped graph. For each pair of amino acids within this graph, if their Euclidean distance is less than 10 Å, a link is created. This process produces local protein structures proportional to the length of each protein and the total number of proteins, resulting in $4,735,677$ local structures from the \texttt{CATH43-S40} dataset. These substructures are encoded into embeddings via a Structure encoder. By varying the parameter K, different structure codebooks are generated using the k-means clustering algorithm.


\subsection{Ablative Homology Retrieval Methods}
\label{sec:app:ablation}

\begin{table}[]
    \caption{Details of the retrieval tools for sequence and structural homologs.}
    \label{tab:retrievalTools}
    \centering
    \resizebox{\linewidth}{!}{
        \begin{tabular}{lll}
        \toprule
        Name & Source & Description \\
        \midrule
        \texttt{EVCouplings} & \url{https://evcouplings.org/} & Selecting sequence homologs by significant numbers.\\
        \texttt{Jackhmmer} & \url{https://www.ebi.ac.uk/Tools/hmmer/search/jackhmmer} & Searching sequence homologs. \\
        \texttt{Foldseek} & \url{https://search.foldseek.com/search} & Searching structure homologs. \\
        \bottomrule
        \end{tabular}
    }
\end{table}

\begin{table}[]
    \caption{Details of the number for homologs of different retrieval tools}
    \label{tab:retrievalNum}
    \centering
    \resizebox{0.8\linewidth}{!}{
        \begin{tabular}{lcccccc}
        \toprule
        Name & File & max. num & avg. num & min. num & median & std. \\
        \midrule
        \texttt{EVCouplings} & a2m & 1,879,223 & 128,297.25 & 44 & 40,154 & 235,938.6 \\
        \texttt{Jackhmmer} & a3m & 26,534 & 8,650.5 & 38 & 8,292 & 6,697.61  \\
        \texttt{Foldseek} & json & 7,773 & 3,529.16 & 0 & 3,606 & 1,985.8 \\
        \bottomrule
        \end{tabular}
    }
\end{table}

We mainly test three methods for homology searching as shown in Table \ref{tab:retrievalTools}. For the 217 assays in ProteinGym, the corresponding \texttt{.a2m} documents can be directly downloaded from \url{https://marks.hms.harvard.edu/proteingym/DMS_msa_files.zip}. We re-searched the homologous sequences of \texttt{ColabFold} and \texttt{Foldseek}, and the specific numbers can be seen in Table \ref{tab:retrievalNum}.

\subsubsection{Evocouplings Alignment File}
For each assay in \textbf{ProteinGym}, we used the curated MSA provided by the official source, where some proteins were retrieved for only the domain regions when the domain was known. The retrieval tool employed was \texttt{Jackhmmer}, a profile HMM-based homology search tool, with $5$ iterations and a bit score ranging from $0.1$ to $0.9$, resulting in $9$ different search groups. The search was conducted on the \texttt{Uniref100} dataset, downloaded on November $24$, $2021$. We then used the \texttt{Evocouplings} library to select the \texttt{.a2m} file with the highest number of significant Evolutionary Couplings (ECs) from the 9 retrieved \texttt{.a2m} files. Since the \texttt{.a2m} files contained lowercase letters and various special characters, we converted lowercase letters to uppercase and replaced special characters with the \texttt{<pad>} token to generate the final alignment file.

\subsubsection{ColabFold Alignment File}
In this study, \texttt{ColabFold} was used to automate the retrieval of multiple sequence alignments (MSA) by querying a local PDB database and generating A3M-formatted files. The \texttt{.a3m} format compresses gaps in the query sequence, improving computational efficiency compared to the \texttt{.a2m} format, which retains gaps for both the query and target sequences. However, the \texttt{.a3m} files output by \texttt{ColabFold} are not fully aligned. To address this, we utilized the \texttt{reformat.pl} script to modify and align the sequences, ensuring they are suitable for downstream analysis. This combined process of automated retrieval and subsequent reformatting optimizes both computational performance and alignment accuracy, streamlining MSA preparation for protein structural and functional studies.

\subsubsection{Foldseek Alignment File}

We automated the retrieval of structure alignment data using the \texttt{Foldseek} \cite{van2024foldseek} API. The process initiates by submitting a query via an \texttt{HTTP POST} request, including various protein structure databases, such as \texttt{afdb50}, \texttt{afdb-proteome}, \texttt{cath50}, \texttt{pdb100}, and others provided in Table \ref{tab:retrievalDatabase}. The query file (\texttt{.pdb}) is uploaded, and the response is monitored in a loop until the search status is marked as \texttt{"COMPLETE"}. During each iteration, the program extracts the result ticket, parses the \texttt{.json} response, and continues polling until successful completion. If an error occurs, the system waits briefly and retries.

Upon receiving the final result, it is processed to extract alignment data. Specifically, the code parses the query and target alignments from the result, removing gaps from the query sequence and adjusting the target sequence accordingly. The final target alignment is padded with gaps to match the query sequence length. The alignments are stored in a dictionary, keyed by target information such as the sequence name, probability, evaluation score, and alignment positions.

\begin{table}[]
    \caption{List of the nine source databases for structural homologs retrieval.}
    \label{tab:retrievalDatabase}
    \centering
    \begin{tabular}{ll}
    \toprule
    Name & Source \\
    \midrule
    \texttt{mgnify\_esm30} & \url{https://www.ebi.ac.uk/metagenomics}\\
    \texttt{afdb50} & \url{https://alphafold.ebi.ac.uk/} \\
    \texttt{afdb-proteome} & \url{https://alphafold.ebi.ac.uk/} \\
    \texttt{cath50} & \url{https://www.cathdb.info/} \\
    \texttt{pdb100} & \url{https://foldseek.steineggerlab.workers.dev/} \\
    \texttt{afdb-swissprot} & \url{https://alphafold.ebi.ac.uk/} \\
    \texttt{gmgcl\_id} & \url{https://gmgc.embl.de/} \\
    \texttt{bfvd} & \url{https://bfvd.steineggerlab.workers.dev/} \\
    \bottomrule
    \end{tabular}
\end{table}
All the datasets can be downloaded in \texttt{Foldseek} format at \url{https://foldseek.steineggerlab.workers.dev/}.

\section{Experimental Setup}
\label{sec:app:config}
The PLM inference is performed on four NVIDIA 3080 GPUs with 10GB VRAM, using \texttt{PyTorch} version 2.4.1. The \texttt{Transformers} library version 4.45.0 is used for loading the PLM. More detailed environment configuration can be found in the environment file \footnote{https://github.com/tyang816/ProtREM/blob/main/environment.yml}. Homology sequence searches are conducted using the conda-maintained \texttt{Jackhmmer} and \texttt{EVCouplings} libraries, with an additional \texttt{plmc} dependency required for \texttt{EVCouplings}. The optimal setting for the retrieval factor $\alpha$ is 0.8. The transformation of homology sequences into matrices is done using the standard ProSST tokenizer \footnote{https://huggingface.co/AI4Protein/ProSST-2048/blob/main/vocab.txt}.

\section{Baseline Methods}
\label{sec:app:baselines}

\begin{table}[]
    \caption{Baseline models.}
    \label{tab:baseline}
    \resizebox{\linewidth}{!}{
        \begin{tabular}{cll}
        \toprule
        Model Name & \multicolumn{1}{c}{Model Type} & \multicolumn{1}{c}{Implementation} \\ 
        \midrule
        SaProt & Hybrid - Structure \& PLM & \url{https://github.com/westlake-repl/SaProt} \\
        Tranception & Hybrid - Alignment \& PLM & \url{https://github.com/OATML-Markslab/Tranception} \\
        GEMME & Alignment-based model & \url{https://hub.docker.com/r/elodielaine/gemme} \\
        ProtSSN & Hybrid - Structure \& PLM & \url{https://github.com/tyang816/ProtSSN} \\
        EVE & Alignment-based model & \url{https://github.com/OATML-Markslab/EVE} \\
        VESPA & Protein language model & \url{https://github.com/Rostlab/VESPA} \\
        Tranception & Hybrid - Alignment \& PLM & \url{https://github.com/OATML-Markslab/Tranception} \\
        MSA Trans & Hybrid - Alignment \& PLM & \url{https://github.com/facebookresearch/esm} \\
        ESM-IF1 & Inverse folding model & \url{https://github.com/facebookresearch/esm} \\
        DeepSequence & Alignment-based model & \url{https://github.com/debbiemarkslab/EVcouplings} \\
        ESM2 & Protein language model & \url{https://github.com/facebookresearch/esm} \\
        ESM-1v & Protein language model & \url{https://github.com/facebookresearch/esm} \\
        MIF-ST & Hybrid - Structure \& PLM & \url{https://github.com/microsoft/protein-sequence-models} \\
        EVmutation & Alignment-based model & \url{https://github.com/debbiemarkslab/EVcouplings} \\
        ESM-1b & Protein language model & \url{https://github.com/facebookresearch/esm} \\ 
        \bottomrule
        \end{tabular}
    }
\end{table}

Our comparison on ProteinGym encompasses a range of state-of-the-art models, including sequence-based models, sequence-structure models, alignment-based models, and inverse folding models. The reproduction and ranking of these models were officially conducted and made publicly available. For the case study comparisons, we specifically focus on the leading sequence-based and sequence-structure models, such as \texttt{SaProt}, \texttt{ProtSSN}, \texttt{ESM2}, and \texttt{MIF-ST}. Detailed implementations of the baselines are provided in Table \ref{tab:baseline}.

\section{Supplementary Results on ProteinGym}
We also evaluated each model's performance on \textbf{ProteinGym} across different MSA depths and taxonomic groups, as shown in Table~\ref{tab:msa_taxon_performance}. ProtREM demonstrated notable improvements across all categories, particularly in scenarios with limited MSA depth and in traditionally challenging categories, such as humans and viruses.

\begin{table}[h!]
\caption{Spearman's $\rho$ correlation of mutation effect prediction (substitution) by zero-shot predictions on \textbf{ProteinGym} of different MSA depth and taxon.}
\centering
\resizebox{\linewidth}{!}{
\begin{tabular}{cccccccccc}
    \toprule
    \multicolumn{1}{l}{\multirow{2}{*}{rank}} & \multirow{2}{*}{model} & \multirow{2}{*}{\textbf{Avg. Spearman}} & \multicolumn{3}{c}{\textbf{MSA Depth}} & \multicolumn{4}{c}{\textbf{Taxon}} \\ \cmidrule(l){4-10} 
    \multicolumn{1}{l}{} &  &  & \multicolumn{1}{l}{Low} & \multicolumn{1}{l}{Medium} & \multicolumn{1}{l}{High} & \multicolumn{1}{l}{Human} & \multicolumn{1}{l}{Eukaryote} & \multicolumn{1}{l}{Prokaryote} & \multicolumn{1}{l}{Virus} \\ 
    \midrule
    \multicolumn{1}{l}{} & ProtREM & \textcolor{red}{\textbf{0.518$\pm$0.000}} & \textcolor{red}{\textbf{0.504}} & \textcolor{red}{\textbf{0.525}} & \textcolor{red}{\textbf{0.572}} & \textcolor{red}{\textbf{0.529}} & \textcolor{red}{\textbf{0.590}} & \textcolor{red}{\textbf{0.538}} & \textcolor{red}{\textbf{0.492}} \\
    \midrule
    1 & SaProt & \textcolor{violet}{\textbf{0.457$\pm$0.000}} & 0.395 & 0.450 & \textcolor{violet}{\textbf{0.545}} & \textcolor{violet}{\textbf{0.476}} & \textcolor{violet}{\textbf{0.523}} & \textcolor{violet}{\textbf{0.524}} & 0.300 \\
    2 & TranceptEVE & \textbf{0.456$\pm$0.008} & \textbf{0.451} & \textbf{0.467} & 0.492 & \textbf{0.471} & 0.498 & 0.473 & \textbf{0.453} \\
    4 & GEMME & 0.455$\pm$0.011 & \textcolor{violet}{\textbf{0.455}} & \textcolor{violet}{\textbf{0.470}} & 0.497 & 0.468 & 0.510 & 0.473 & \textcolor{violet}{\textbf{0.469}} \\
    6 & ProtSSN & 0.449$\pm$0.006 & 0.401 & 0.458 & 0.522 & 0.469 & \textbf{0.518} & 0.505 & 0.356 \\
    10 & EVE & 0.439$\pm$0.010 & 0.425 & 0.453 & 0.481 & 0.453 & 0.487 & 0.468 & 0.428 \\
    14 & VESPA & 0.436$\pm$0.007 & 0.427 & 0.455 & 0.484 & 0.438 & 0.492 & 0.490 & 0.432 \\
    15 & Tranception & 0.434$\pm$0.008 & 0.432 & 0.438 & 0.473 & 0.453 & 0.483 & 0.431 & 0.432 \\
    16 & MSA Trans & 0.434$\pm$0.008 & 0.393 & 0.435 & 0.473 & 0.427 & 0.491 & 0.451 & 0.390 \\
    22 & ESM-if1 & 0.422$\pm$0.011 & 0.300 & 0.431 & \textbf{0.544} & 0.415 & 0.497 & \textbf{0.507} & 0.374 \\
    24 & DeepSequence & 0.419$\pm$0.012 & 0.383 & 0.428 & 0.473 & 0.442 & 0.469 & 0.460 & 0.344 \\
    26 & ESM2 & 0.414$\pm$0.011 & 0.335 & 0.406 & 0.515 & 0.456 & 0.471 & 0.476 & 0.238 \\
    28 & ESM-1v & 0.407$\pm$0.012 & 0.326 & 0.405 & 0.499 & 0.456 & 0.446 & 0.432 & 0.279 \\
    32 & MIF-ST & 0.400$\pm$0.009 & 0.376 & 0.403 & 0.455 & 0.398 & 0.415 & 0.462 & 0.397 \\
    33 & EVmutation & 0.395$\pm$0.008 & 0.403 & 0.421 & 0.410 & 0.409 & 0.444 & 0.422 & 0.388 \\
    34 & ESM-1b & 0.394$\pm$0.009 & 0.350 & 0.398 & 0.482 & 0.434 & 0.475 & 0.455 & 0.241 \\
    \bottomrule\\[-2.5mm]
    \multicolumn{10}{l}{$\dagger$ The top three scores are highlighted by \textbf{\textcolor{red}{First}}, \textbf{\textcolor{violet}{Second}}, and \textbf{Third}.}
\end{tabular}
}
\label{tab:msa_taxon_performance}
\end{table}

We held out 10\% of mutants from each assay as a validation set to evaluate the performance of different retrieval methods and the generalizability of retrieval factors. The detailed results are presented in Table~\ref{tab:retrieval_ratios}. Notably, the optimal performance is achieved when using the official \texttt{.a2m} homologous sequence with $\alpha=0.8$.

\begin{table}
\caption{Performance of retrieval ratios for different methods on 10\% samples of each assay of ProteinGym.}
\centering
    \begin{tabular}{lccccccccc}
    \toprule
     Retrieval & \textbf{0.1} & \textbf{0.2} & \textbf{0.3} & \textbf{0.4} & \textbf{0.5} & \textbf{0.6} & \textbf{0.7} & \textbf{0.8} & \textbf{0.9} \\
    \midrule
    \texttt{EVCouplings} & 0.516 & 0.518 & 0.526 & 0.521 & 0.526 & 0.532 & 0.529 & 0.541 & 0.529 \\
    \texttt{ColabFold} & 0.516 & 0.517 & 0.525 & 0.519 & 0.522 & 0.527 & 0.524 & 0.534 & 0.521 \\
    \texttt{Foldseek} & 0.515 & 0.517 & 0.524 & 0.518 & 0.521 & 0.527 & 0.522 & 0.533 & 0.526 \\
    \texttt{EVCouplings}+\texttt{Foldseek} & 0.516 & 0.519 & 0.528 & 0.525 & 0.532 & 0.536 & 0.527 & 0.506 & 0.430 \\
    \bottomrule
    \end{tabular}
\label{tab:retrieval_ratios}
\end{table}

\begin{figure}
    \centering
    \includegraphics[width=1\linewidth]{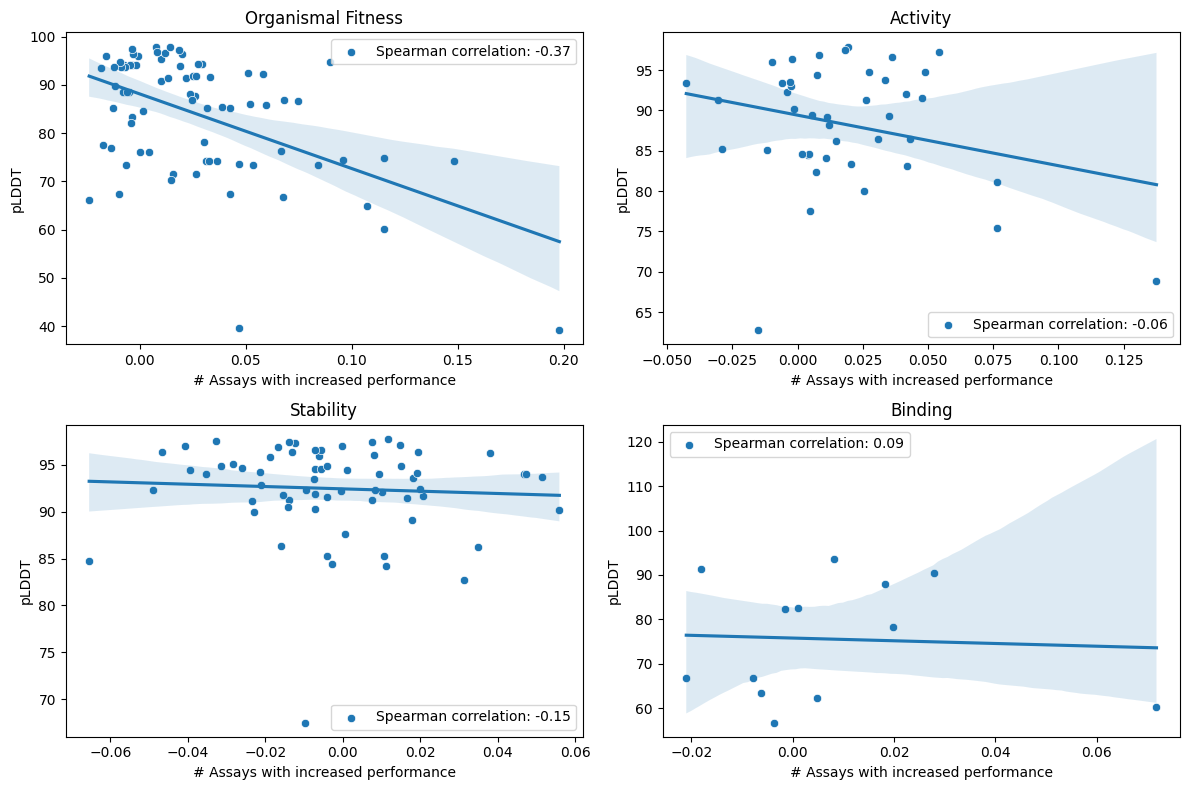}
    \caption{Spearman score difference between with and without retrieval by pLDDT.}
    \label{fig:plddt_func}
\end{figure}

\begin{figure}
    \centering
    \includegraphics[width=1\linewidth]{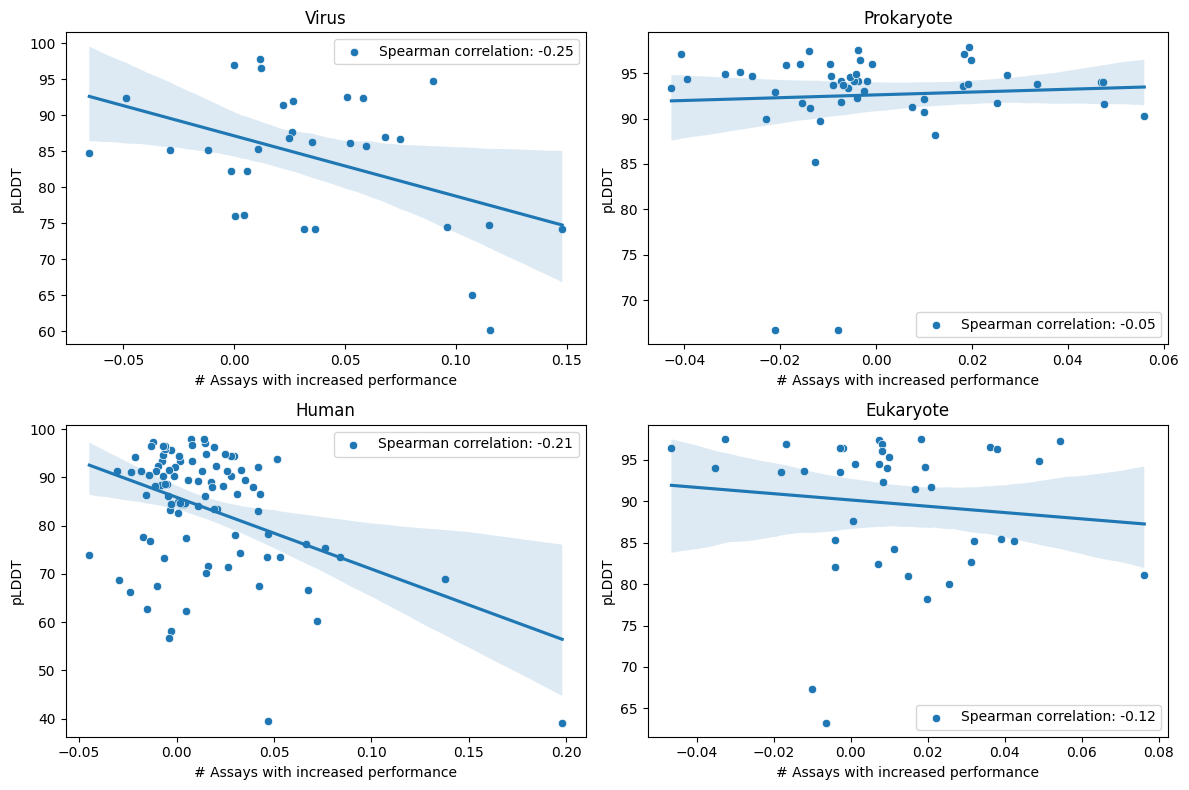}
    \caption{Spearman score difference between with and without retrieval by taxon.}
    \label{fig:plddt_taxon}
\end{figure}

\begin{figure}
    \centering
    \includegraphics[width=0.83\linewidth]{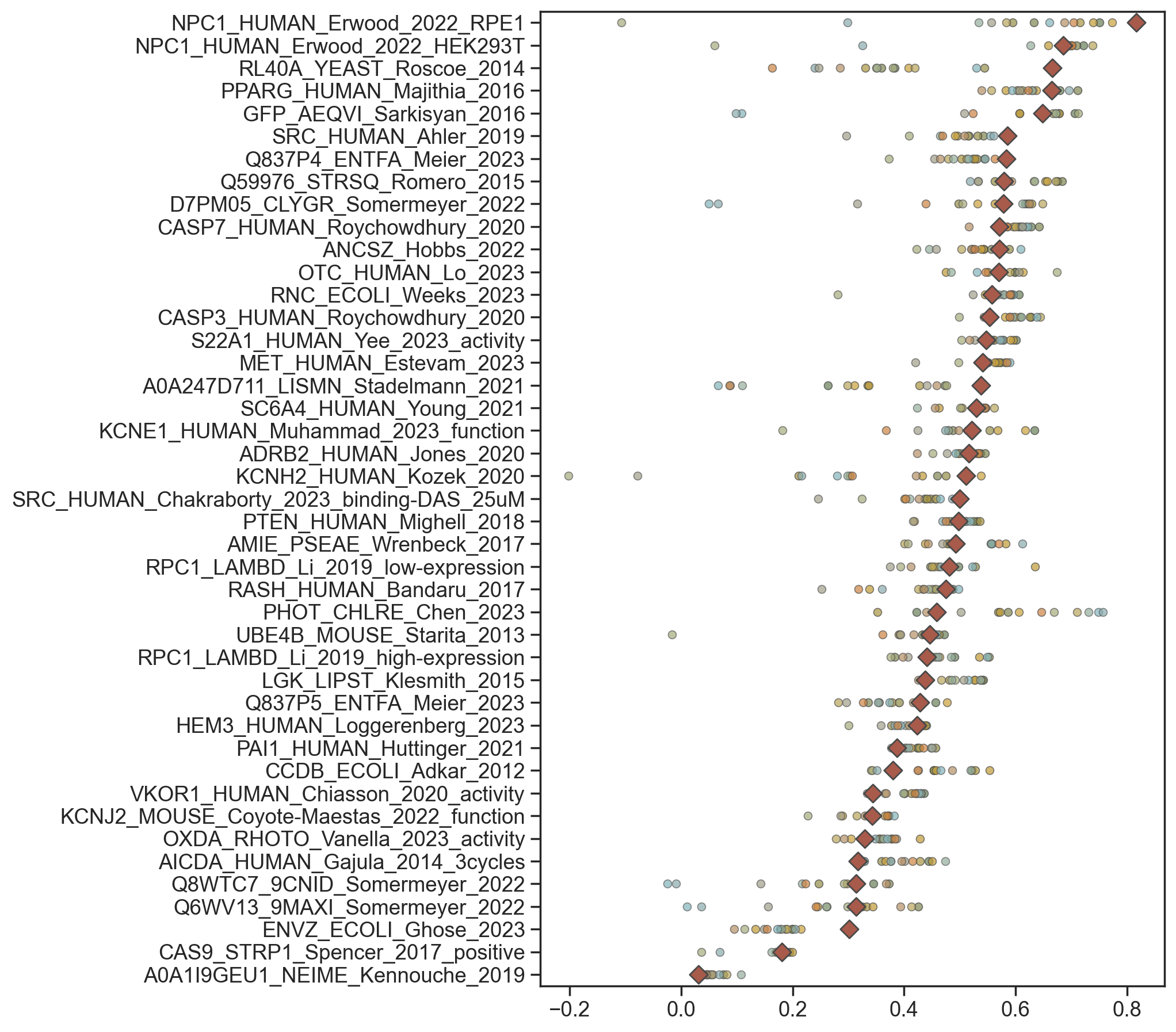}
    \includegraphics[width=0.15\linewidth,trim = 0cm -10cm 0cm 0cm]{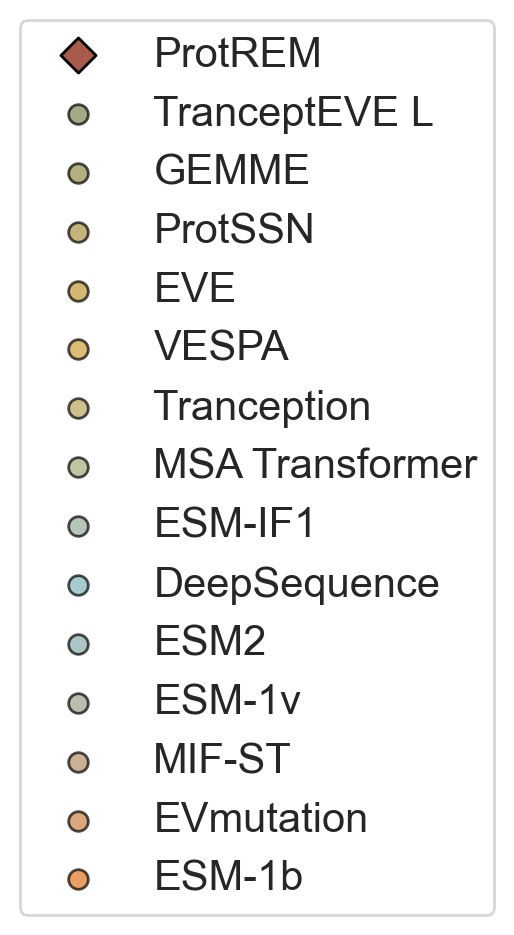}
    \caption{Activity}
    \label{fig:assay_activity}
\end{figure}

\begin{figure}
    \centering
    \includegraphics[width=0.83\linewidth]{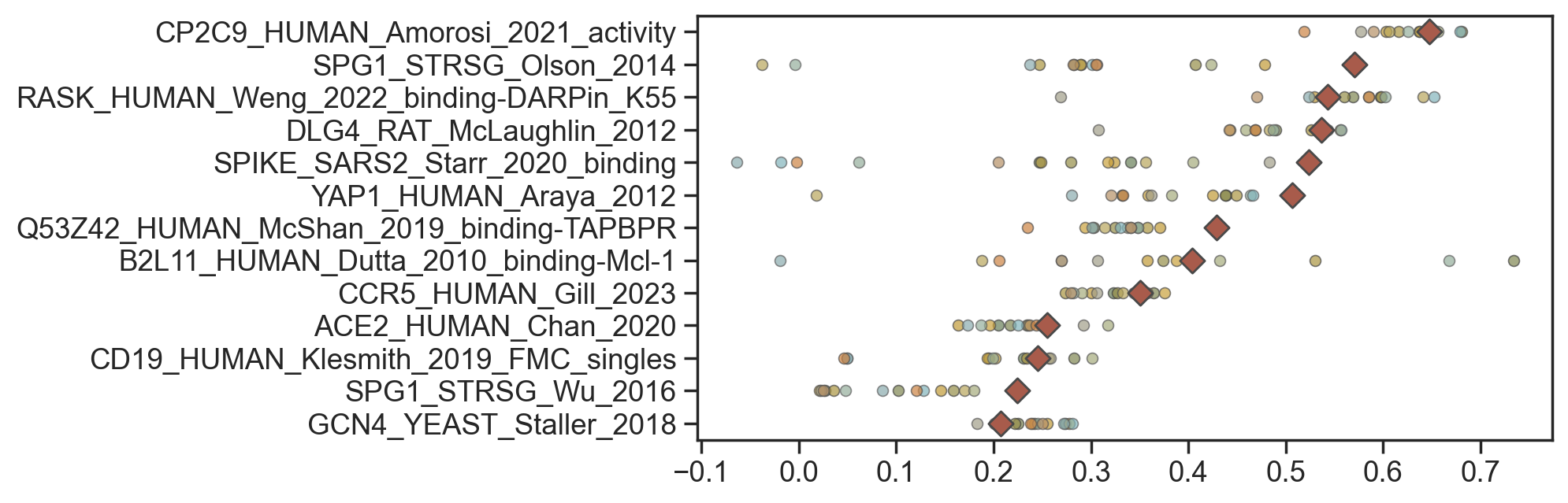}
    \includegraphics[width=0.15\linewidth,trim = 0cm 0cm 0cm 0cm]{figure/legend.png}
    \caption{Binding}
    \label{fig:assay_binding}
\end{figure}

\begin{figure}
    \centering
    \includegraphics[width=0.83\linewidth]{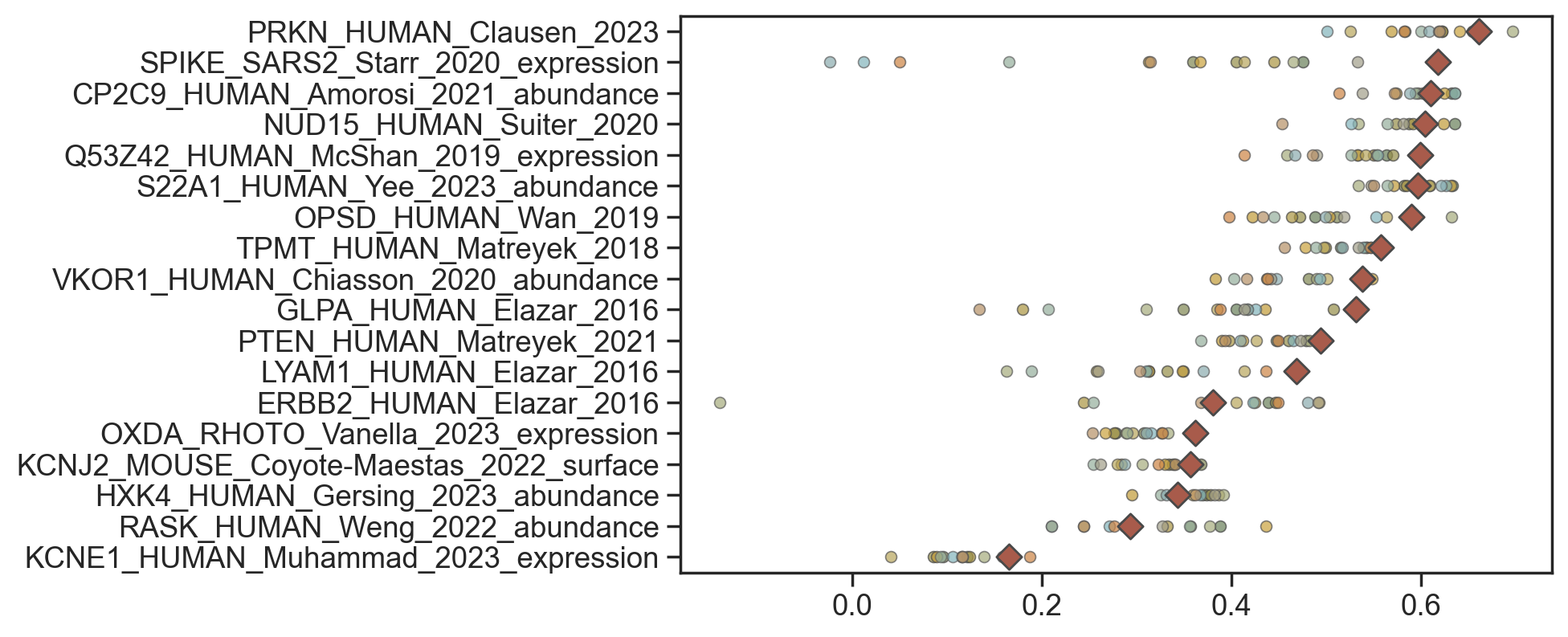}
    \includegraphics[width=0.15\linewidth,trim = 0cm 0cm 0cm 0cm]{figure/legend.png}
    \caption{Expression}
    \label{fig:assay_expression}
\end{figure}

\begin{figure}
    \centering
    \includegraphics[width=0.83\linewidth]{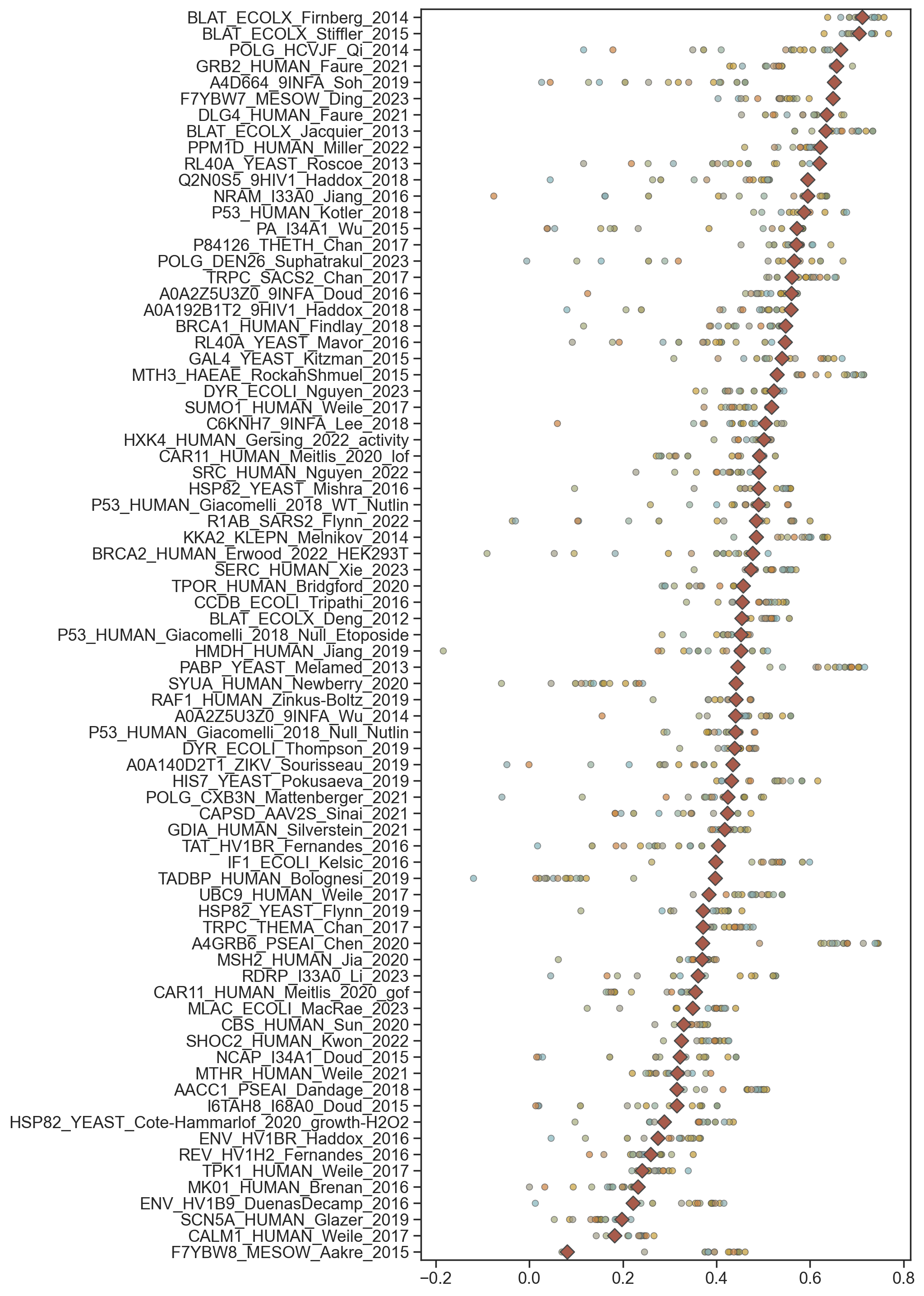}
    \includegraphics[width=0.15\linewidth,trim = 0cm -15cm 0cm 0cm]{figure/legend.png}
    \caption{Organismal Fitness}
    \label{fig:assay_organ}
\end{figure}

\begin{figure}
    \centering
    \includegraphics[width=0.83\linewidth]{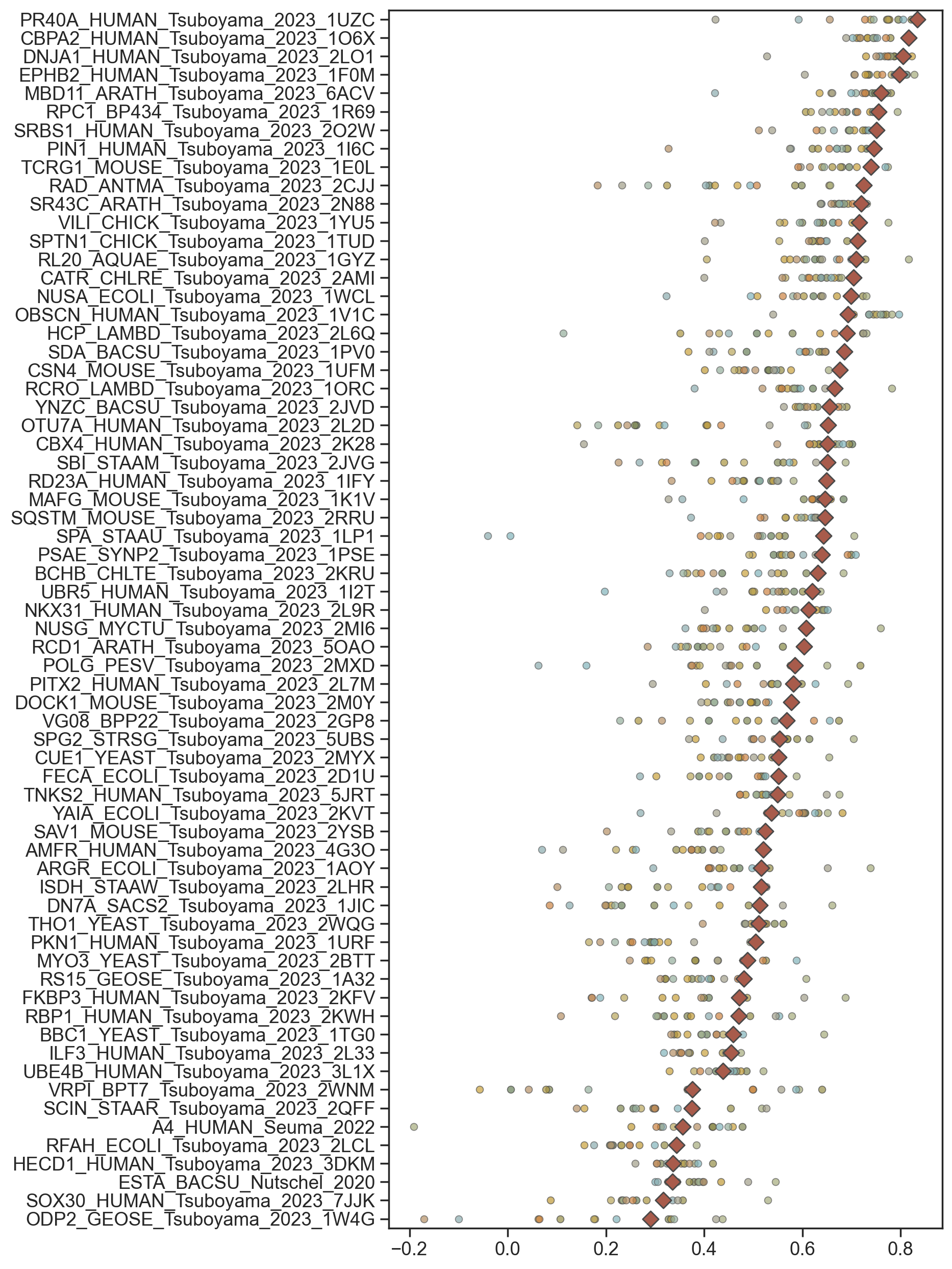}
    \includegraphics[width=0.15\linewidth,trim = 0cm -18cm 0cm 0cm]{figure/legend.png}
    \caption{Stability}
    \label{fig:assay_stability}
\end{figure}

With homologous information and predicted protein structures being simultaneously input into the model, we aim to determine whether the explicit addition of homologous information can compensate for deficiencies in protein structure quality. As shown in Fig.~\ref{fig:plddt_func} and \ref{fig:plddt_taxon}, a generally negative correlation between \texttt{pLDDT} and score improvement is observed across categories, except for Binding (likely due to a limited number of data points). This indicates that the lower the structural quality, the more effective the homologous information becomes. The detailed Spearman scores can be found in Figs~\ref{fig:assay_activity}-\ref{fig:assay_stability}.

\section{Data Processing for Case Studies}
\label{sec:app:caseStudyData}
We extracted the experimental data of $31$ VHH antibody mutants from \cite{kang2024vhh}, covering 1-4 site mutations for binding affinity and alkali resistance. The relevant experimental methods are provided in Appendix~\ref{sec:app:expWet}. We used the average EC50 values from three repeated experiments as the quantitative metric for both assays.

The sequence of wild-type VHH antibody we used in this study is listed in Table~\ref{tab:phi29sequences}. For the structures, we use \texttt{AlphaFold3} server \cite{abramson2024alphafold3} instead of \texttt{ColabFold} to fold the amino acid sequence to get high-quality structures. For the homology alignment files, we use the \texttt{UniRef100} database downloaded on October $2024$. The remaining processing methods and parameters remain unchanged (see Appendix \ref{sec:app:config}).





\begin{table}
\caption{The wild-type sequences of phi29 DNAP AA (C3) and VHH.}
\label{tab:phi29sequences}
\centering
\resizebox{0.8\linewidth}{!}{
\begin{tabular}{cc} \\ 
 \toprule 
 Protein & Sequence\\  
 \midrule 
         C3 & 
         MKHMPRKRYSCDFETTTKVEDCRVWAYGYMNIEDHSEYKIGNSLDE\\&
         FMAWALKVQADLYFHNLKFDGAFIINWLERNGFKWSADGLPNTYNT\\&
         IISRMGQWYMIDICLGYKGKRKIHTVIYDSLKKLPFPVKKIAKDFKLT\\&
         VLKGDIDYHKERPVGYKITPEEYAYIKNDIQIIAEALLIQFKQGLDRMT\\&
         AGSDSLKDFKDIITTKKFKKVFPTLSLELDKKVRYAYRGGFTWLNDRF\\&
         KEKEIGEGMVFDVNSLYPAQMYSRLLPYGEPIVFEGKYVWDEDYPLH\\&
         IQHIRCEFELKEGYIPTIQIKRSRFYKGNEYLKSSGGEIADLWLSNVDLE\\&
         LMKEHYDLYNVEYISGLKFKATTGLFKDFIDKWTYIKTTSEGAIKQLA\\&
         KLMLNSLYGKFASNPDVTGKVPYLKENGALGFRLGEEETKDPVYTPM\\&
         GVFITAWARYTTITAAQACYDRIIYCDTDSIHLTGTEIPDVIKDKVDPKK\\&
         LGYWAHESTFKRAKYLRPKTYIQDIYMKEVDGELVEGSPDDYTDIKFS\\&
         VKCAGMTDKIKKEVTFENFKVGFSRKMKPKPVQVPGGVVLVDDTFTIK \\ 
         \midrule
         VHH & 
         MQVQLVESGGGLAQAGGSLRLSCAVSGMPEFARAMGWFRQAPGKERE\\&
         LLAAIEGIGATTYYADSVKGRFTISRDDAANTVLLQMNSLKPDDTAVY\\&
         YCAAAFSVTIPTRARHWVDWGPGTLVTVSSDDDDKSGGGGSHHHHHH \\

         \bottomrule 
    \end{tabular}
    }
\end{table}

\section{Experimental Methods}
\label{sec:app:expWet}

\paragraph{Binding Affinity of VHH Antibody}
Ninety-six-well plates were coated with growth hormone protein at a density of 5 ng per well at 4 $^{\circ}$C overnight. The plates were washed with 1 $\times$ PBS$^{\prime}$T three times. Following blocking with 1\% BSA in 1 $\times$ PBS at 25$^{\circ}$C for 2 hours. After washing three times with 1 $\times$ PBS$^{\prime}$T, the plates were incubated with serial dilutions of VHH proteins 100 $\mu$L per well (1:2, 1:4, 1:8, 1:16, 1:32, 1:64, 1:128, 1:256, 1:512, 1:1024, and 1:2048) for 1 hour at 25$^{\circ}$C. After washing three times with 1 $\times$ PBS$\prime$T, 100 $\mu$L/well HRP-labeled Goat Anti-Mouse IgG(H+L) (1:5000) were added and incubated at 25$^{\prime}$C for 1h. The plates were washed with 1 $\times$ PBS$^{\prime}$T four times, a total of 100 $\mu$L/well L 3, 3$^{\prime}$,5, 5$^{\prime}$-tetramethylbenzidine was added and incubated at 25 $^{\prime}$C for 15 minutes in the dark. Finally, 100 $\mu$L/well 2 M H\textsubscript{2}SO\textsubscript{4} was added to stop the reaction and absorbance was measured at 450 nm (TECAN, Swiss).
The log(agonist) versus response -- Variable slope (four parameters) curves were analyzed to calculate EC50 which determines the stability of VHH after alkaline treatment.

\paragraph{Alkaline pH stability test of VHH Antibody (ELISA)}
The VHH antibodies (1.5 mg/mL) were treated an equal volume of 0.3 M or 0.5 M NaOH for 24 hours, followed by pH adjustment using the same volume of 0.5 M HCl, the supernatant was collected after centrifugation.

Ninety-six-well plates were coated with growth hormone protein at a density of 5 ng/well at 4 $^{\circ}$C overnight. The plates were washed with 1 $\times$ PBS$^{\prime}$T three times. Following blocking with 1\% BSA in 1 $\times$ PBS at 25$^{\circ}$C for 2 hours. After washing three times with 1 $\times$ PBS$^{\prime}$T, the plates were incubated with serial dilutions of VHH proteins 100 $\mu$L/well (1:2, 1:4, 1:8, 1:16, 1:32, 1:64, 1:128, 1:256, 1:512, 1:1024, and 1:2048) for 1 hour at 25$^{\circ}$C. After washing three times with 1 × PBS’T, 100 $\mu$L/well HRP-labeled Goat Anti-Mouse IgG(H+L) (1:5000) were added and incubated at 25$^{\circ}$C for 1 hour. The plates were washed with 1 $\times$ PBS$^{\prime}$T four times, a total of 100 $\mu$L/well L 3, 3$^{\prime}$,5, 5$^{\prime}$-tetramethylbenzidine was added and incubated at 25 $^{\circ}$C for 15 minutes in the dark. Finally, 100 $\mu$L/well 2 M H\textsubscript{2}SO\textsubscript{4} was added to stop the reaction and absorbance was measured at 450 nm (TECAN, Swiss).

The log(agonist) vs. response -- Variable slope (four parameters) curves were analyzed to calculate EC50 which determines the stability of VHH after alkaline treatment.

\paragraph{Construction of phi29 DNAP expression plasmid }
The gene of phi29 DNA polymerase and its variants genes were synthesized by Sangon Biotech (Shanghai, China) after codon-optimized. These genes were subsequently cloned into the pET28(a) plasmid,, which features an N-terminal His-tag for protein purification.

\paragraph{Protein expression of phi29 DNAP}
The expression plasmid was transformed into \textit{E. coli} BL21(DE3). A 30 mL seed culture was initially cultivated at 37 °C in LB medium supplemented with 50 µg/ml kanamycin, and this culture was then transferred to 500 mL of LB medium containing the same concentration of kanamycin in a shaker flask. The cultures were incubated at 37 $^{\circ}$C until the OD600 reached 1.0, and protein expression was then induced by the addition of isopropyl-D-thiogalactopyranoside (IPTG) to a final concentration of 0.5 mM, followed by incubation for 16 h at 25 $^{\circ}$C.

\paragraph{Protein purification of phi29 DNAP}
Cells were harvested by centrifugation for 30 minutes at 4,000 rpm, and the resulting pellets were collected for subsequent purification. The cell pellets were then resuspended in lysis buffer (25 mM Tris-HCl, 500 mM NaCl, pH 7.4) and disrupted via ultrasonication (Scientz, China). The lysates were centrifuged at 12,000 rpm for 30 minutes at 4 $^{\circ}$C, and the supernatants were subjected to Ni-NTA affinity purification using an elution buffer composed of 25 mM Tris-HCl, 500 mM NaCl, and 250 mM imidazole (pH 7.4). Following purification, the protein was desalted with lysis buffer through ultrafiltration. The protein-containing fractions were then flash-frozen at -20 $^{\circ}$C in a storage buffer consisting of 25 mM Tris-HCl (pH 7.4), 200 mM NaCl, and 20\% glycerol.

\end{document}